\documentclass[letterpaper]{article} 
\usepackage{aaai25}  
\usepackage{times}  
\usepackage{helvet}  
\usepackage{courier}  
\usepackage[hyphens]{url}  
\usepackage{graphicx} 
\usepackage{natbib}  
\usepackage{caption} 
\usepackage{algorithm}
\usepackage{algorithmic}

\usepackage{newfloat}
\usepackage{listings}
\DeclareCaptionStyle{ruled}{labelfont=normalfont,labelsep=colon,strut=off} 
\floatstyle{ruled}
\newfloat{listing}{tb}{lst}{}
\floatname{listing}{Listing}
\newcommand{\halftablewidth}{0.45\textwidth}
\usepackage{graphicx}
\usepackage{makebox}
\usepackage{rotating}
\usepackage{makecell}
\usepackage{booktabs}  
\usepackage{adjustbox} 
\usepackage{float}
\usepackage{amsmath} 
\usepackage{amsfonts}
\usepackage{xcolor}
\usepackage{subcaption}
\usepackage{colortbl}
\definecolor{lightgreen}{rgb}{0.886, 0.941, 0.851}
\definecolor{red}{rgb}{1, 0, 0}
\definecolor{blue}{rgb}{0, 0, 1}
\definecolor{lightgrey}{rgb}{0.8627, 0.8627, 0.8627}
\definecolor{D95326}{rgb}{0.851,0.325,0.149}
\title{Do Not DeepFake Me: Privacy-Preserving Neural 3D Head Reconstruction
Without Sensitive Images}
\author{
    Jiayi Kong\textsuperscript{\rm 1},
    Xurui Song\textsuperscript{\rm 1},
    Shuo Huai,\textsuperscript{\rm 2},
    Baixin Xu\textsuperscript{\rm 1},
    Jun Luo\textsuperscript{\rm 1},
    Ying He\textsuperscript{\rm 1}\thanks{Corresponding author}  
}
\affiliations{
    \textsuperscript{\rm 1}S-Lab, Nanyang Technological University, Singapore\\

    \textsuperscript{\rm 2}College of Computing and Data Science, Nanyang Technological University, Singapore

   \{JIAYI006, SONG0257\}@e.ntu.edu.sg, \{shuo.huai, baixin.xu, junluo, YHe\}@ntu.edu.sg
}

\usepackage{bibentry}

\begin{document}

\maketitle

\begin{abstract}
While 3D head reconstruction is widely used for modeling, existing neural reconstruction approaches rely on high-resolution multi-view images, posing notable privacy issues. Individuals are particularly sensitive to facial features, and facial image leakage can lead to many malicious activities, such as unauthorized tracking and deepfake. In contrast, geometric data is less susceptible to misuse due to its complex processing requirements, and absence of facial texture features. In this paper, we propose a novel two-stage 3D facial reconstruction method aimed at avoiding exposure to sensitive facial information while preserving detailed geometric accuracy. Our approach first uses non-sensitive rear-head images for initial geometry and then refines this geometry using processed privacy-removed gradient images. Extensive experiments show that the resulting geometry is comparable to methods using full images, while the process is resistant to DeepFake applications and facial recognition (FR) systems, thereby proving its effectiveness in privacy protection.
\end{abstract}

\vspace{-0.18in}
\section{Introduction}
\label{sec:intro}

In the rapidly evolving digital era, 3D facial reconstruction technology has become an indispensable component in fields such as modeling, virtual reality, and digital entertainment. Recent advancements, particularly the emergence of techniques like Neural Radiance Fields (NeRF)~\cite{mildenhall2020nerf} and Signed Distance Function (SDF)-based methods (NeuS/VolSDF)~\cite{wang2021neus,yariv2021volume}, have significantly enhanced the accuracy of reconstruction, heralding a revolutionary progression in this domain. However, these methods often depend on multi-view, high-resolution facial images. While such images provide substantial convenience and precision in reconstruction, they raise significant privacy concerns due to the sensitive nature of the detailed facial information they contain. This reliance on sensitive images inevitably compromises privacy.

Extensive research has consistently demonstrated that individuals exhibit a markedly higher sensitivity to facial features compared to other body parts~\cite{zebrowitz2008social}. Even minor facial imperfections, such as acne, scars, or uneven skin tone, can exert considerable psychological effects~\cite{hamler2022skin,mekeres2023importance}. The potential misuse of facial information raises substantial concerns, including risks of social discrimination, psychological distress, and other negative consequences~\cite{abrams2020african}. Furthermore, the necessity for protecting facial privacy extends beyond individual sensitivities to include the broader risks associated with the leakage of facial images~\cite{ciftci2023my}. Current image-based facial recognition (FR) systems have reached a level of sophistication that enables the extraction and analysis of extensive information from facial images~\cite{kortli2020face}. In the event of facial image leakage, this data could be exploited for unauthorized digital tracking, surveillance, or other malicious purposes~\cite{hill2022secretive}. Moreover, facial images provide the precise details necessary for creating DeepFake~\cite{chadha2021deepfake}, further amplifying privacy risks.

In contrast, geometric data is generally less sensitive compared to facial images. 
Geometric data alone lacks the detailed facial textures necessary for impersonation and forgery, such as those used in DeepFake technology~\cite{westerlund2019emergence}. This absence of detailed textures significantly diminishes the risk of misuse. Additionally, collecting, processing, and analyzing 3D data is more complex and resource-intensive~\cite{uy2021joint}, which hinders efficient tracking and recognition. The technology for handling 3D data is less developed and less widespread than that for 2D images~\cite{guo20233d}, further reducing the risk of unauthorized monitoring and tracking.
Thus, while facial geometry also reveals some information, it is much harder to exploit in DeepFake and FR systems, resulting in fewer malicious effects than the facial images used in current face reconstruction methods.
This phenomenon highlights the urgent need to improve current methods and explore high-quality 3D reconstruction techniques that do not depend on high-risk facial images.

Existing privacy protection methods, such as image blurring \cite{jiang2023dartblur} or facial masking \cite{sun2018natural}, while somewhat effective in 2D image processing, face numerous challenges in 3D reconstruction tasks.  
mm3DFace~\cite{xie2023mm3dface} is one of the approaches that address facial privacy in 3D reconstruction by using mmWave signals to extract geometric features, thus avoiding facial images and offering some level of privacy protection. It can track 68 facial landmarks, which is useful for expression analysis. However, it cannot capture geometric details which is necessary for nuanced applications.
Currently, there is a lack of a comprehensive end-to-end strategy capable of reconstructing detailed facial geometry without relying on sensitive facial images. 

To address this challenge, we propose an innovative 3D neural reconstruction method without dependence on sensitive facial images, representing the first systematic attempt to address privacy protection issues in the field of neural surface reconstruction. Our method employs a two-stage reconstruction process using two types of images: privacy-neutral images (e.q. rear-head images) and privacy-protected images, which are initially facial images processed to remove sensitive information. We first utilize privacy-neutral images to establish the basic geometric structure and then use privacy-protected images to refine and perfect the 3D reconstruction. This approach completely eliminates the reliance on sensitive facial RGB images while still enabling the reconstruction of stable and detailed geometric head models, thus avoiding the risk of sensitive image leakage.

Our research strikes a balance between the requirements of head reconstruction and user privacy protection, 
offering a reliable and secure head geometry reconstruction solution for various application scenarios. 
Our main contributions can be summarized as follows:

\begin{itemize}
\item 
We introduce a novel end-to-end neural reconstruction pipeline that effectively protects facial privacy in 3D modeling. Our method eliminates the exposure of sensitive information while maintaining high-quality geometric reconstruction, thereby broadening NeRF's applications in privacy-sensitive scenarios and laying a foundation for privacy-protected advancements.
\item 
We develop a unique two-stage training process that balances reconstruction stability and detail fidelity. This approach first demonstrates that gradient images can effectively contribute to geometric reconstruction, preventing reliance on sensitive facial images. 
\item 
Our method significantly improves the security and effectiveness of neural reconstruction in sensitive contexts. Using images that lack detailed facial textures, protects facial privacy and renders them ineffective for common exploitation techniques such as DeepFake creation and facial recognition.
\end{itemize}

\section{Related Work}
\label{sec:formatting}
\paragraph{Human head models.} 3D Morphable Models (3DMM)~\cite{blanz2023morphable} leverage principal component analysis to represent facial. However, this method falls short in capturing details such as wrinkles, the interior of the mouth, and hair, so it may not fully satisfy appearance requirements. i3DMM~\cite{yenamandra2021i3dmm} introduces an implicit function to model both the geometry and appearance of the human head with various attributes like shape, expression, and hairstyle. Both approaches rely on 3D data for their modeling. Recent advancements in Neural Radiance Fields (NeRF)~\cite{mildenhall2020nerf,barron2021mipnerf} have excelled in novel view synthesis due to their compact and powerful representation capability, relying solely on a set of multi-view images. Consequently, numerous works~\cite{gafni2021dynamic,park2021nerfies} have yielded detailed geometry~\cite{zheng2022avatar, Xu_2023_ICCV} and achieved a photo-realistic appearance~\cite{zheng2023neuface, KirschsteinQGWN23} in modeling heads. 
\paragraph{Neural implicit functions.}
Sign distance fields (SDF) \cite{park2019deepsdf} and occupancy fields~\cite{mescheder2019occupancy}, showcase their representative ability over explicit representations, i.e. mesh, and point cloud. DVR~\cite{niemeyer2020differentiable} and IDR~\cite{yariv2020multiview} focus on differentiating the surface rendering pipeline based on multi-view images. They incorporate corresponding masks to distinguish objects from the background during the training process. NeuS~\cite{wang2021neus} and VolSDF~\cite{yariv2021volume} refine the geometry reconstruction of NeRF by introducing a scheme that converts SDF to density, enhancing the surface representation in NeRF. Recent approaches~\cite{rosu2023permutosdf, wang2022hf, wang2023pet} combine volume rendering with multi-scale hashing in Instant-NGP~\cite{muller2022instant} and displacement fields to learn detailed surfaces through implicit functions from multi-view images. 

\paragraph{Facial privacy protection.}We summarize three categories primary approaches. The first involves an algorithm based on adversarial generation networks, aimed at deceiving unauthorized facial recognition by generating fake images highly similar to the original ones~\cite{li2019privacy}. However, this method is primarily targeted at public social platforms~\cite{ciftci2023my} or specific facial recognition (FR) systems that provide data for learning~\cite{DBLP:conf/iclr/CherepanovaGFDD21}. 
The second category involves the use of cryptographic techniques. Cryptographic techniques, including homomorphic encryption~\cite{huang2020instahide}, secure multiparty computation~\cite{ma2019lightweight}, and other encryption primitives~\cite{kou2021efficient}, are used to encrypt original images securely. These approaches introduce higher latency and computational costs. The third group of methods focuses on obfuscation techniques. These methods encompass actions such as implementing blurring~\cite{jiang2023dartblur}, introducing noise~\cite{zhang2021adaptive}, applying masking~\cite{seneviratne2022does}, utilizing filtering~\cite{zhou2020personal}, and employing image transformation~\cite{wang2021gender}. While practical, these methods irreversibly degrade image quality and may fail in subsequent tasks without a robust end-to-end solution. mm3DFace~\cite{xie2023mm3dface} uses mmWave signals for privacy protection, but its goal is not to reconstruct detailed geometry.

\vspace{-0.1in}
\section{Method}
\begin{figure*}
  \centering
  \includegraphics[width=0.9\textwidth]{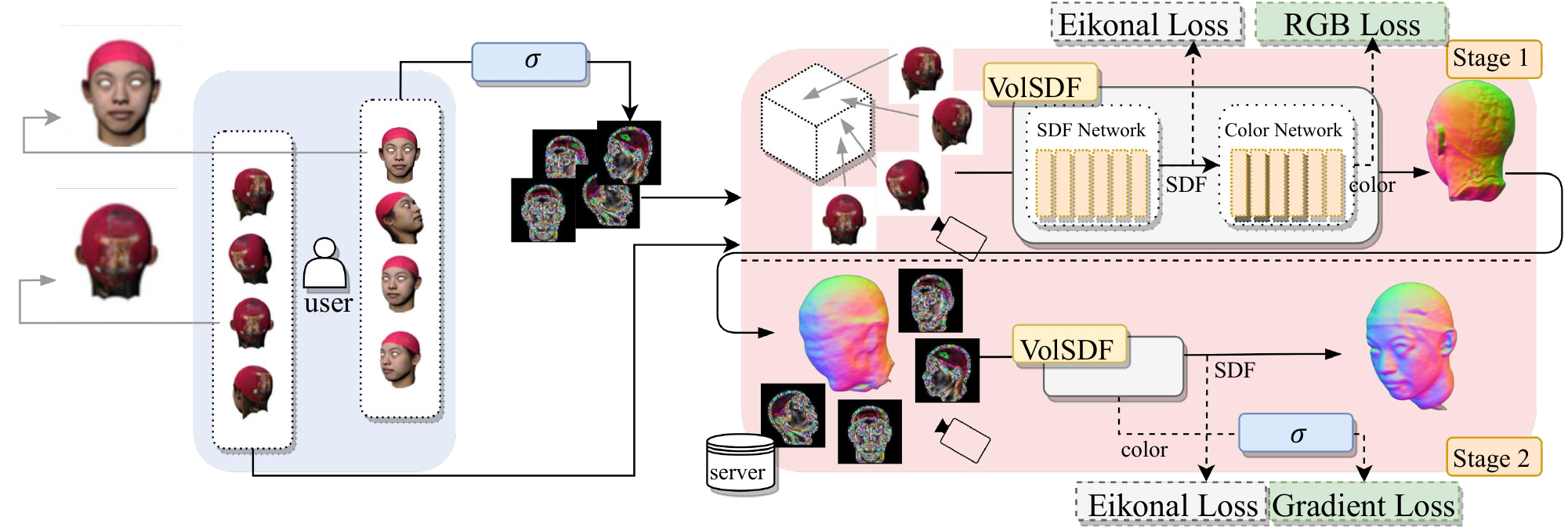}
\vspace{-0.1in}
  \caption{ \textbf{Pipeline}. We simulate the data flow process where our model reconstructs the human head geometry, without depending on sensitive images. Users can select photos that require privacy protection through a general operator denoted by $\sigma$. Following this, privacy-neutral images and privacy-protected images are used for geometric reconstruction. The reconstruction process occurs in two stages: stage 1 uses privacy-neutral images to establish the foundational geometry, while stage 2 employs image gradient information to refine the facial geometry further. }
  \label{fig:pipeline}\vspace{-0.1in}
\end{figure*}
\subsection{Preliminaries}\label{3.1}
\textbf{Neural volume rendering.} As demonstrated by NeRF~\cite{mildenhall2020nerf}, it characterizes a 3D scene by employing volume density and color fields. Given known camera poses and ray directions, we conduct sampling along the rays, predicting both the color $\mathbf{c}_i$ and density $\sigma_i$ at each sample point $\mathbf{x}_i$ using Multi-Layer Perceptron (MLPs). The volume rendering process entails the integrating color radiance across the sampled points along the ray. Consequently,  we approximate the rendered color for a specific pixel as:
\begin{equation}\label{eq:1}
\hat{\mathbf{c}} (\mathbf{o},\mathbf{d})=\sum_{i=1}^{N}\omega _{i}\mathbf{c} _{i}.
\end{equation}
In this context, we define the distance of each sample point from the camera center as $t_i$ and introduce $\delta_i$ to represent the distance between adjacent sample points: $\delta_i = t_{i+1} - t_i$. Furthermore, we use $\alpha_i$ to quantify the opacity of the i-th ray segment, computed as $\alpha_i = 1 - \exp(-\sigma_i \delta_i)$. The term $T_{i}= {\textstyle \prod_{j=1}^{i-1}(1-\alpha _{j})} $ denotes the accumulated transmittance, indicating the proportion of light that reaches the camera, and $\omega _{i}$ is defined as \(\omega _{i}=T_{i}\alpha _{i}\) to determine the in Eq.~\eqref{eq:1}.
We train the network using the color loss between the rendered images and input images:
\begin{equation}\label{eq:2}
\mathcal{L}_{\text{rgb}}=\left \|  \hat{\mathbf{c}}-\mathbf{c} \right \| _{1}.
\end{equation}
\paragraph{Volume rendering of SDF.} One of the most common surface representations is the SDF, which precisely describes an object's geometry. An SDF for a 3D object with a watertight surface is a function \( f : \mathbb{R}^3 \rightarrow \mathbb{R} \). This function takes any point\( \mathbf{x} \in \mathbb{R}^3 \) and provides the signed distance between \( \mathbf{x} \) and the closest point on the surface. Importantly, the zero-level set of the SDF corresponds to the object's surface \( S \):
\begin{equation}\label{eq:3}
S = \{ \mathbf{x} \in \mathbb{R}^3 \,|\, f(\mathbf{x}) = 0 \}.
\end{equation}
Recent contributions have been centered around neural volume rendering based on SDF representation~\cite{wang2021neus,yariv2021volume}. These methods utilize MLPs to implicitly represent a 3D scene by predicting the SDF and color $\mathbf{c}_i$ relative to the viewpoint. This differs from the initial approach presented in~\cite{mildenhall2020nerf}, which focused on predicting color and density. The extraction of the zero-level surface of the SDF in these methods results in a more reasonable geometric representation, showcasing significant effectiveness in reconstructing objects with smooth geometry.

If we treat a function as an SDF, it must adhere to the requirement of differentiability, ensuring that the modulus of the gradient remains constant in accordance with the eikonal equation. Therefore, we incorporate the eikonal loss into the final SDF predictions to ensure that the optimized $f(\mathbf{x}_i)$ conforms to a valid SDF:
\begin{equation}\label{eq:4}
\mathcal{L}_{\text{eik}}=\frac{1}{N}\sum_{i=1}^{N}(\| \nabla f({\mathbf{x}_{i}}) \| _2-1)^{2},
\end{equation}
where $N$ represents the total number of sampled points. Given our use of a network architecture for SDF prediction, computing gradients within a continuous field becomes both feasible and straightforward.

\subsection{Two-stage Training}\label{3.2}
In our assumptions, the reconstruction process should neither access nor alter sensitive facial images, ensuring that the algorithm maintains practical utility for geometric reconstruction while fully adhering to privacy requirements. To implement this approach, all images used for reconstruction must be non-sensitive and are categorized into two types: \textit{privacy-neutral} and \textit{privacy-protected}. Neutral pictures, which do not contain facial information, such as those captured from the back of the head, are directly utilized in the training process. Privacy-protected images, initially taken from frontal perspectives, are processed through a specialized operator to ensure they meet our privacy goals before being uploaded by the user and used in the second stage.

In our proposed method, we employ a two-stage training framework for geometric reconstruction in Figure~\ref{fig:pipeline}. In the first stage, we train a neural radiance field using neutral images, which enables the reconstruction of essential low-frequency information.  
The success of this stage depends on the quantity of privacy-neutral data users provide. While this information might have been deemed less valuable for facial geometry learning in the past, it proves to be beneficial in our method. Unless unavoidable circumstances require the use of templates, as discussed in the section \textbf{Optimization}, the privacy-neutral images provided by the user can be fully utilized. This enhances the geometric accuracy in the first stage of reconstruction, bringing it closer to the user's geometric features, and ultimately contributing to the overall quality of the head reconstruction in the second stage.

In the second stage, we utilize privacy-protected data uploaded by users for the reconstruction process. Our supervision focuses solely on color variation information inherent in the original images. We intentionally refrain from utilizing full RGB information to mitigate the risk of privacy exposure.
While the privacy-protected images appear visually unfriendly and blurry, they contain valuable color variation details crucial for refining geometric intricacies. Building on the foundation established in the first stage, we train for new frontal face viewpoints. We compute gradients of the rendered images, transforming them into multi-view color gradient modulus through a general operator $\sigma$. These modulus are then compared to the facial privacy-protected gradient information obtained from the user after passing through the operator. To optimize this stage, we introduce a novel loss equation, the gradient loss $\mathcal{L}_{\text{grad}}$, to guide the learning of geometric information in the second stage:
\begin{equation}\label{eq:5}
\mathcal{L}_{\text{grad}}=\big \| {\| \hat{\mathbf{g}} \|_2 }-\left \|\mathbf{g} \right \|_2 \big\|_{1},
\end{equation}
where $\mathbf{g}$ and $\hat{\mathbf{g}}$ are color gradients of the original image and the predicted gradient information, respectively.
We calculate the gradient of the color in both  $x-$ and $y-$ directions and obtain their modulus as follows:
\begin{equation}\label{eq:7}
\left \|\mathbf{g} \right \|_2 =\left \|\frac{\partial {\mathbf{c}} }{\partial x} \right \|_2+\left \|\frac{\partial {\mathbf{c}} }{\partial y}  \right \|_2.
\end{equation}\
In practical implementation, this operator can take the form of an edge extraction operator, such as the Sobel operator. Its design aims for versatility, enabling users to choose from a range of operators $\sigma$ depending on their specific privacy requirements and to adjust operator parameters to ensure the irreproducibility of the reconstruction. This flexibility empowers users to acquire data with different levels of protection.
During this stage, the absence of RGB supervision may lead to inaccuracies in MLP color predictions. Nevertheless, our method effectively recovers facial geometric details even when radiance information appears unreliable. 

The privacy-preserving approach focuses on three key aspects: irreversibility by retaining only gradient magnitudes, color multiplicity where different images can map to the same gradient, and perceptual indistinction to keep the processed data visually indistinguishable from the original. Detailed discussions are in the supplementary material.
\subsection{Optimization}\label{sec:optimization}
\paragraph{Template.}Templates offer an effective solution when users cannot capture photos without privacy-sensitive information, such as handheld devices or other constraints. In such cases, user-uploaded images need processing for privacy protection. To address this challenge, we propose using a neutral head template. This enables users to recover head geometry even when privacy-neutral images are scarce, facilitating effective geometric reconstruction. While template usage is not mandatory, it enhances our model's compatibility with various data inputs, making it more versatile for a smoother user experience.
\vspace{-0.1in}
\paragraph{Regularization.}In our network structure, based on the neural radiance fields pipeline, we make specific optimizations to enhance geometric representation. Our primary goal is not the final rendering appearance but the recovery of fine geometric details for downstream tasks. To achieve this, we apply regularization constraints to the color rendering network. These constraints encourage the network to prioritize learning geometric intricacies. In previous work~\cite{rosu2023permutosdf}, a color regularization constraint was proposed which introduced trainable bounds for each layer, effectively constraining the expression of MLP layers using knowledge from Lipschitz continuous networks. In the specific network structure, each MLP layer is reformulated as $y=\sigma (\widehat{W}_{i}\mathbf{x}+b_{i})$, and $ \widehat{W}_{i} =a(W_{i},\text{softplus}(m_{i}))$, where $a$ normalizes the weight matrix. During the privacy-protected stage, we apply color regularization by constraining the product of Lipschitz constants, $m_{i}$, for each layer:
\begin{equation}\label{eq:8}
\text{softplus}(m_{i})=\ln_{}{(1+e^{m_{i}}) }.
\end{equation}

In the training process for stage one, we do not use this regularization term. In the second stage of training, we introduce constraints on the rendering network as an additional loss:
\begin{equation}\label{eq:9}
\mathcal{L}_{\text{lip}}= {\textstyle \prod_{i=1}^{l}}\text{softplus}(m_{i}).
\end{equation}
Putting it all together, our training loss is as follows:
\begin{equation}\label{eq:10}
\mathcal{L}= \lambda_{1}\mathcal{L}_{\text{rgb}}+\lambda_{2}\mathcal{L}_{\text{eik}}+\lambda_{3}\mathcal{L}_{\text{lip}}+\lambda_{4}\mathcal{L}_{\text{grad}}.
\end{equation}
In Stage 1, we set $\lambda_{3}=\lambda_{4}=0$ to disable gradients, and set $\lambda_{1}=0$ in Stage 2 to activate it.

\section{Experiments}
\vspace{-0.1in}
\subsection{Setup}
\begin{figure}[t] \centering
    \makebox[0.01\textwidth]{}
    \makebox[0.104\textwidth]{\small ID 08}
    \makebox[0.104\textwidth]{\small ID 00}
    \makebox[0.104\textwidth]{\small ID 03}
    \\
    \includegraphics[width=0.104\textwidth]{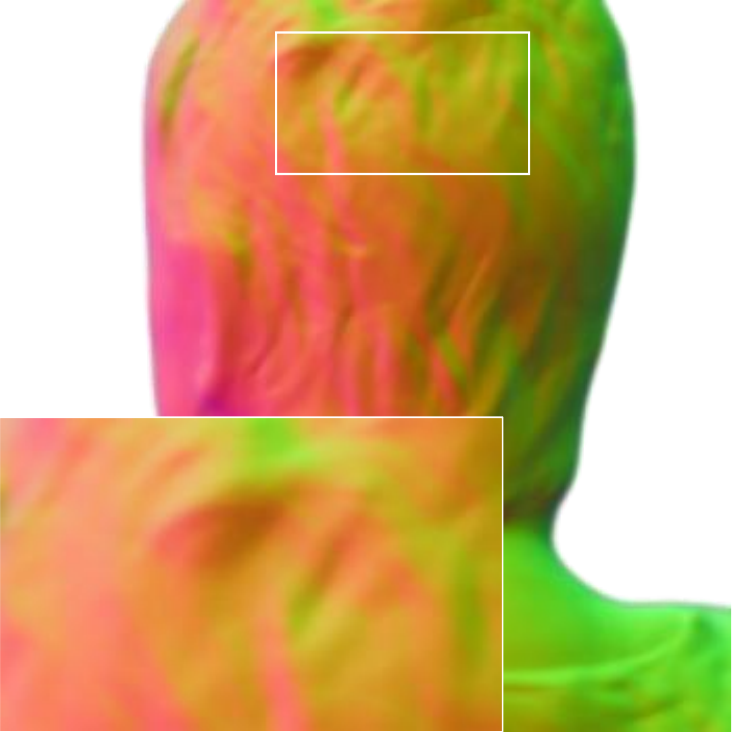}
    \includegraphics[width=0.104\textwidth]{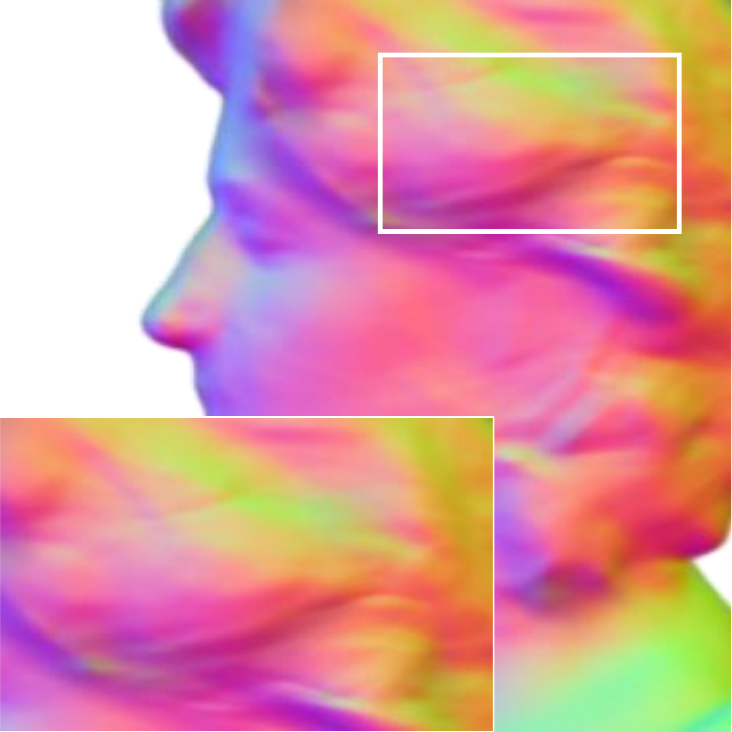}
    \includegraphics[width=0.104\textwidth]{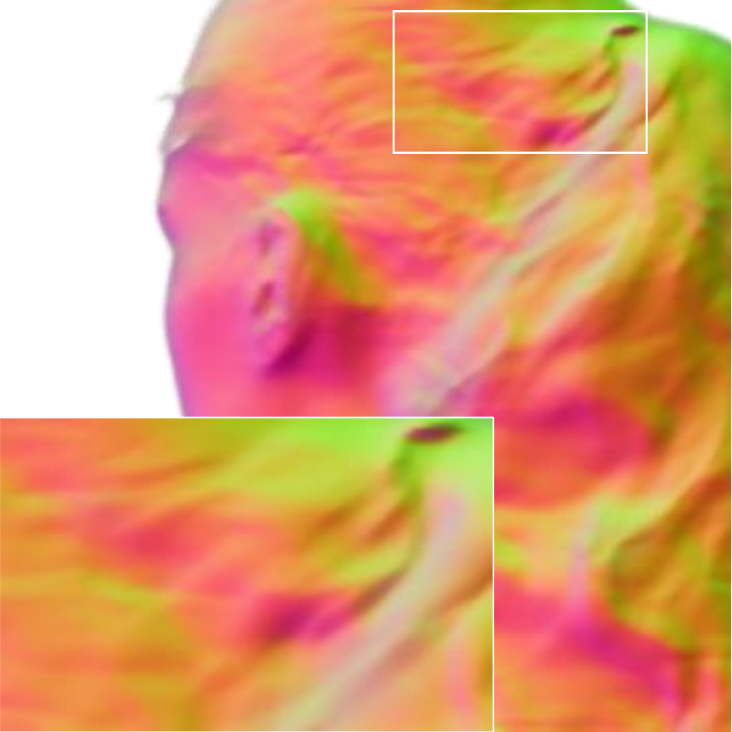} 
    \\
    \
    \makebox[0.01\textwidth]{}
    \makebox[0.104\textwidth]{\small angry}
    \makebox[0.104\textwidth]{\small shock}
    \makebox[0.1042\textwidth]{\small smile}
    \\
    \includegraphics[trim=1cm 2 3cm 1cm, clip,width=0.104\textwidth]{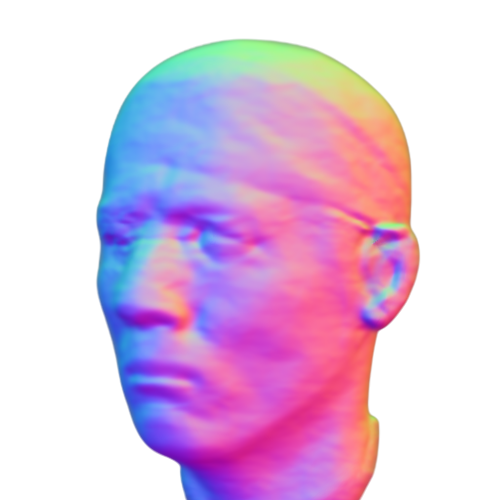}
    \includegraphics[trim=1cm 2 3cm 1cm, clip,width=0.104\textwidth]{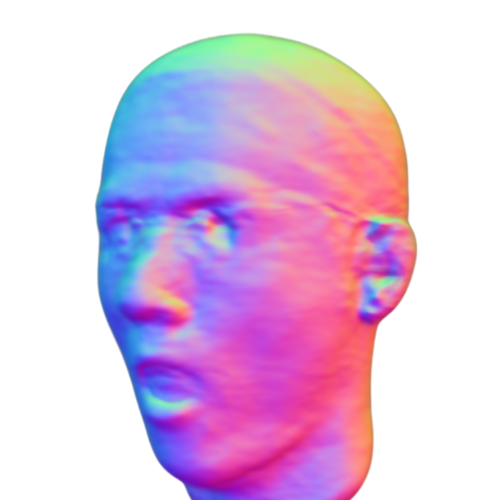}
     \includegraphics[trim=1cm 2 3cm 1cm, clip,width=0.104\textwidth]{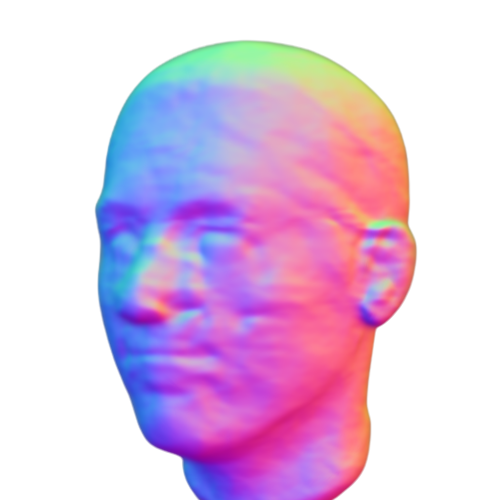} \vspace{-0.1in}
    \caption{In terms of reconstruction quality, even with 64-resolution images, we are still able to capture many geometric details, including hair and expressions.}\vspace{-0.1in}
    \label{fig:hair}
\end{figure}

\paragraph{Datasets.}
In our experiments, we utilize two representative datasets: FaceScape~\cite{yang2020FaceScape} and High-Fidelity 3D Head (H3DS)~\cite{ramon2021h3d}. Each dataset includes 30 to 36 RGB images per identity at $64 \times 64$ pixels, with lower resolution and increased blurring chosen to enhance privacy protection. FaceScape offers textured 3D face data for various subjects and expressions, while H3DS, captured in real-world scenarios, provides headshot data across different countries, ethnic backgrounds, and lighting conditions. We randomly sampled 10 identities from each dataset and treated all images with visible facial features as sensitive images. The selection of images for training is adaptable to user privacy preferences, as detailed in the supplementary material. In our experiments with the FaceScape dataset, we utilize 10 images that are inherently privacy-neutral and process 20 images as privacy-protected. Similarly, for the H3DS dataset, 16 images are inherently privacy-neutral, and 20 images are processed as privacy-protected. While our primary analysis focuses on these two datasets, we also test additional facial datasets to broaden our validation in the supplementary. These experiments demonstrate the adaptability and robustness of our method in various conditions.
\begin{figure}[H] \centering
    \makebox[0.01\textwidth]{}
    \makebox[0.088\textwidth]{\scriptsize GT image}
    \makebox[0.088\textwidth]{\scriptsize Stage 1}
    \makebox[0.088\textwidth]{\scriptsize Stage 2}
    \makebox[0.088\textwidth]{\scriptsize VolSDF (full img)}
    \makebox[0.088\textwidth]{\scriptsize GT mesh}
    \\
    \raisebox{0.4\height}{\makebox[0.01\textwidth]{\rotatebox{90}{\makecell{\scriptsize ID 393}}}}
    \includegraphics[trim=1cm 0cm 3cm 1cm,  clip,scale=1.2,width=0.088\textwidth]{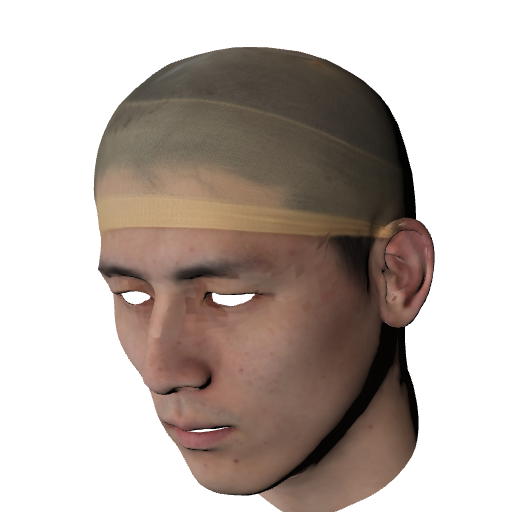}
    \includegraphics[trim=1cm 0cm 3cm 1cm,  clip,scale=1.2,width=0.088\textwidth]{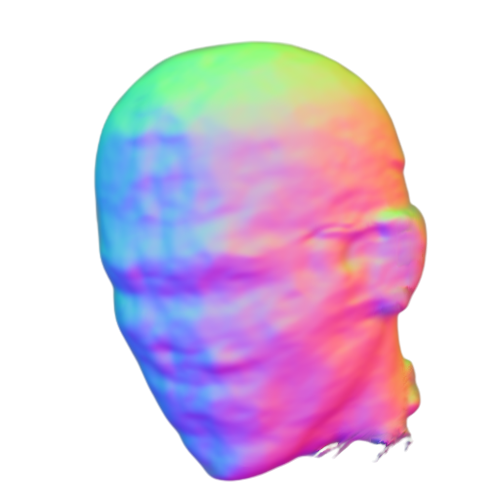}
    \includegraphics[trim=1cm 0cm 3cm 1cm,  clip,scale=1.2,width=0.088\textwidth]{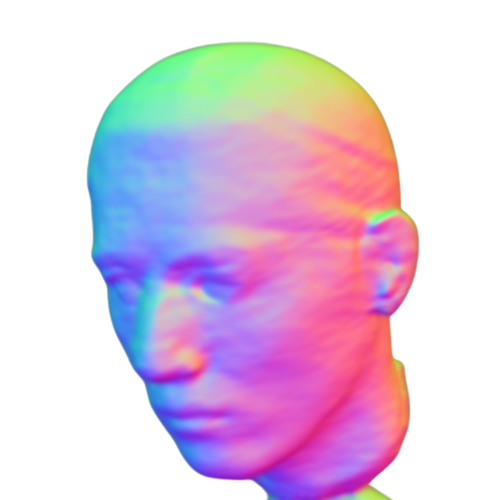}
    \includegraphics[trim=1cm 0cm 3cm 1cm, clip,scale=1.2,width=0.088\textwidth]{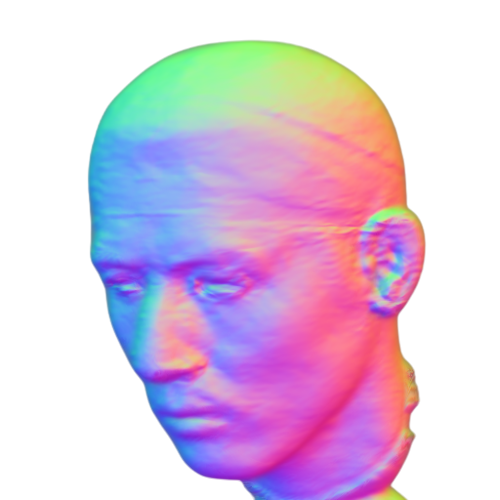}
    \includegraphics[trim=1cm 0cm 3cm 1cm, clip,scale=1,width=0.088\textwidth]{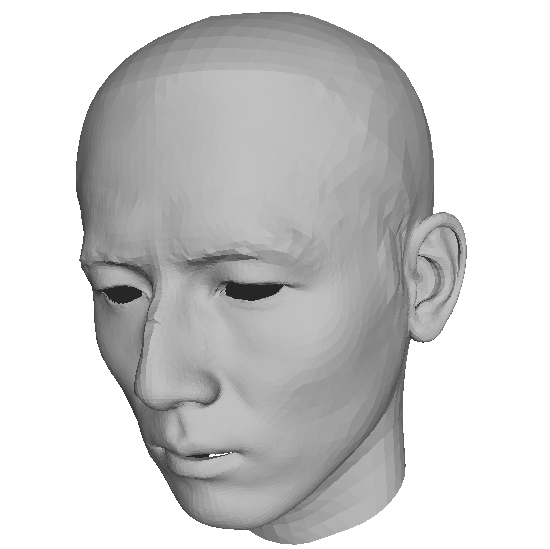 }
    \\
    \raisebox{0.4\height}{\makebox[0.01\textwidth]{\rotatebox{90}{\makecell{\scriptsize ID 421}}}}
    \includegraphics[trim=1cm 0cm 3cm 1cm,  clip,scale=1.2,width=0.088\textwidth]{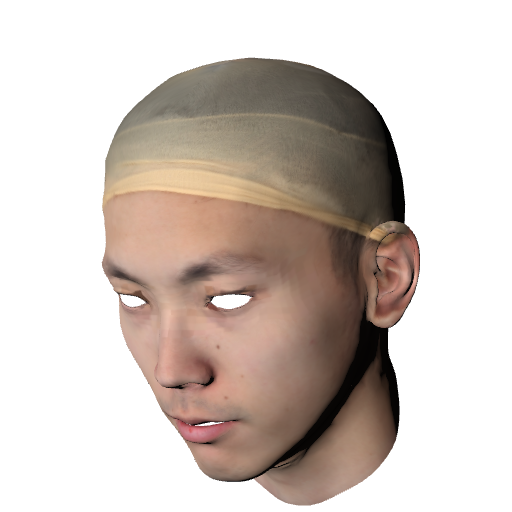}
    \includegraphics[trim=1cm 0cm 3cm 1cm,  clip,scale=1.2,width=0.088\textwidth]{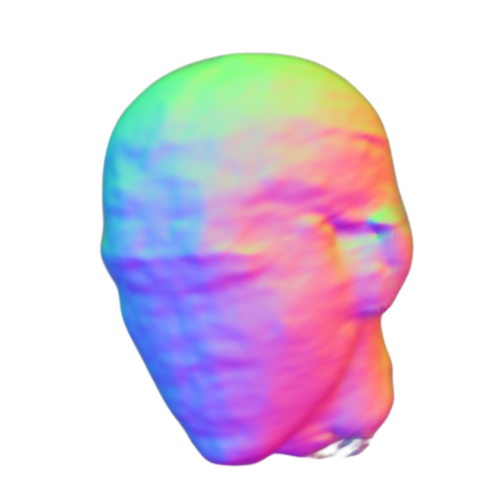}
    \includegraphics[trim=1cm 0cm 3cm 1cm,  clip,scale=1.2,width=0.088\textwidth]{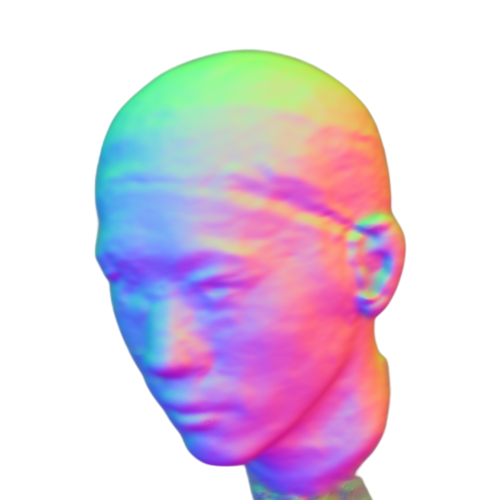}
    \includegraphics[trim=1cm 0cm 3cm 1cm,  clip,scale=1.2,width=0.088\textwidth]{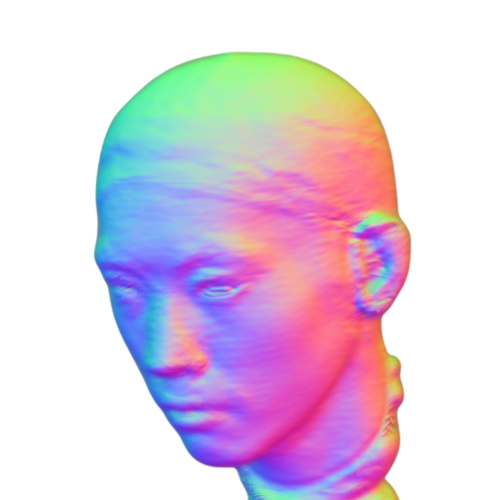}
    \includegraphics[trim=1cm 0cm 3cm 1cm, clip,scale=1.2,width=0.088\textwidth]{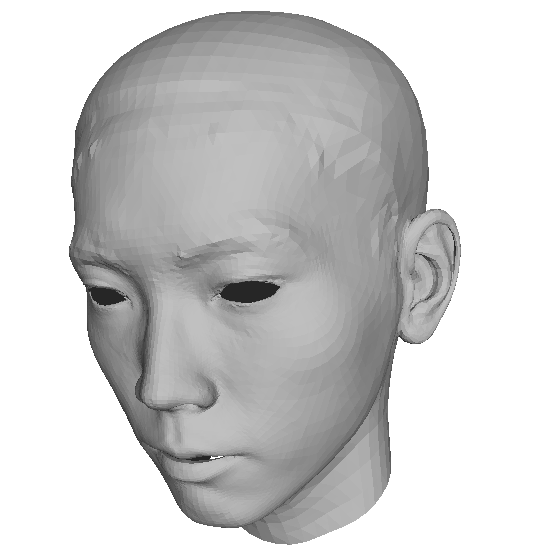 }
    \\
    \raisebox{0.4\height}{\makebox[0.01\textwidth]{\rotatebox{90}{\makecell{\scriptsize ID 395}}}}
    \includegraphics[trim=1cm 0cm 3cm 1cm, clip,scale=1.2,width=0.088\textwidth]{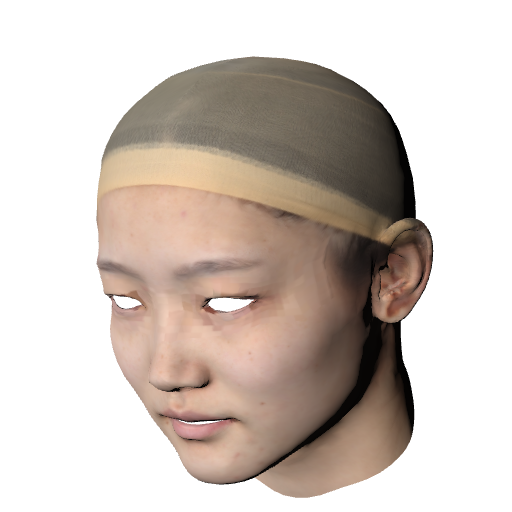}
    \includegraphics[trim=1cm 0cm 3cm 1cm,  clip,scale=1.2,width=0.088\textwidth]{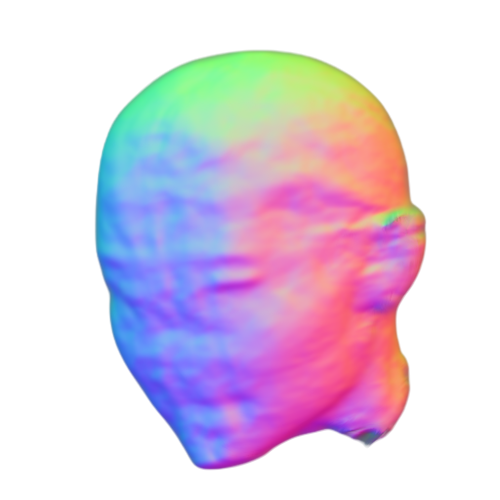}
    \includegraphics[trim=1cm 0cm 3cm 1cm,  clip,scale=1.2,width=0.088\textwidth]{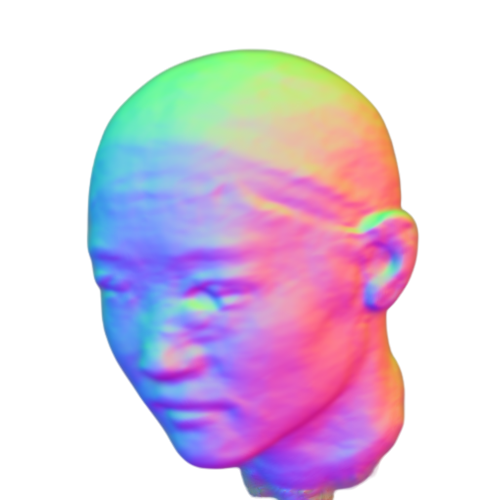}
    \includegraphics[trim=1cm 0cm 3cm 1cm,  clip,scale=1.2,width=0.088\textwidth]{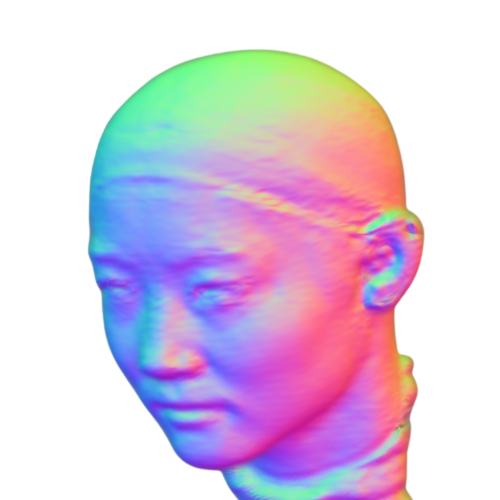}
    \includegraphics[trim=1cm 0cm 3cm 1cm, clip,scale=1.2,width=0.088\textwidth]{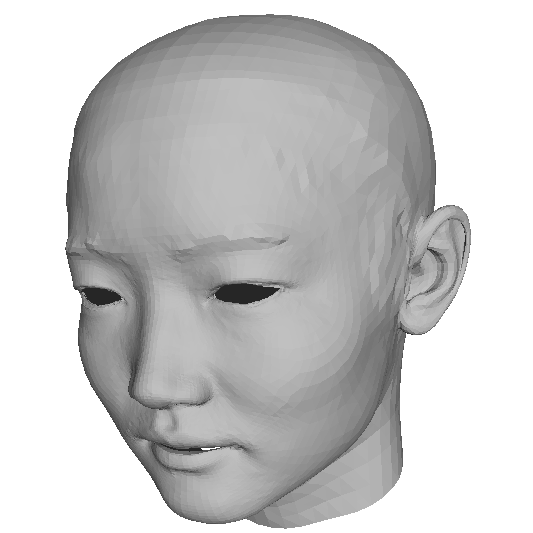 }
    \\
    \raisebox{0.4\height}{\makebox[0.01\textwidth]{\rotatebox{90}{\makecell{\scriptsize ID 00}}}}
    \includegraphics[width=0.088\textwidth]{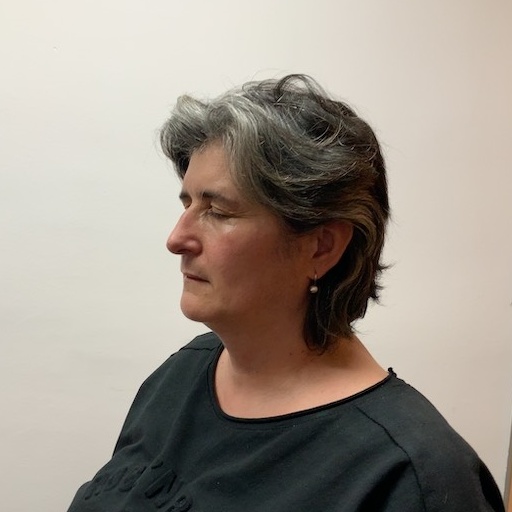}
    \includegraphics[width=0.088\textwidth]{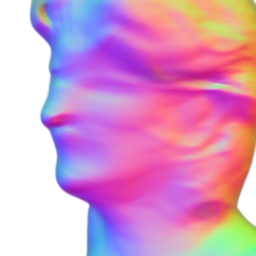}
    \includegraphics[width=0.088\textwidth]{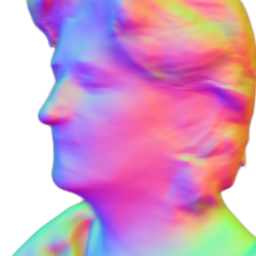}
    \includegraphics[width=0.088\textwidth]{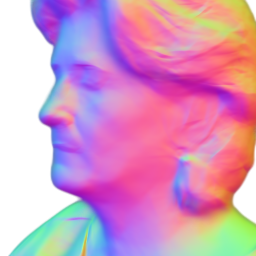}
     \includegraphics[width=0.088\textwidth]{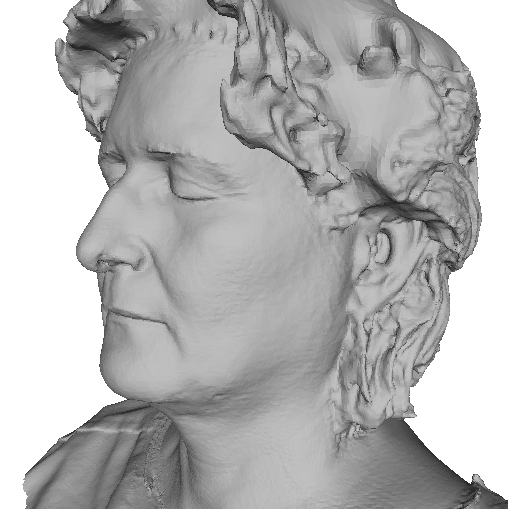}
    \\
     \raisebox{0.4\height}{\makebox[0.01\textwidth]{\rotatebox{90}{\makecell{\scriptsize ID 09}}}}
     \includegraphics[width=0.088\textwidth]{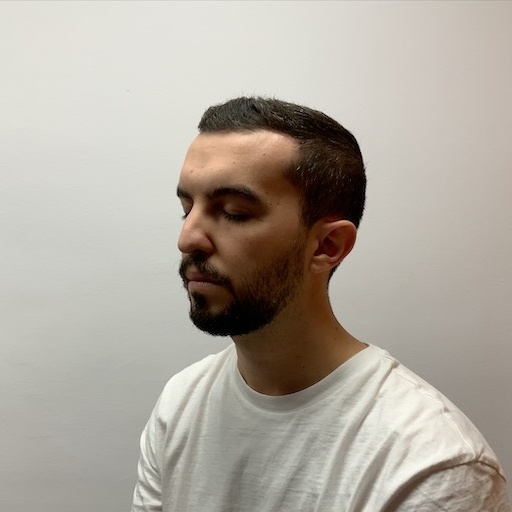}
    \includegraphics[width=0.088\textwidth]{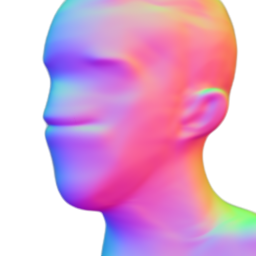}
   \includegraphics[width=0.088\textwidth]{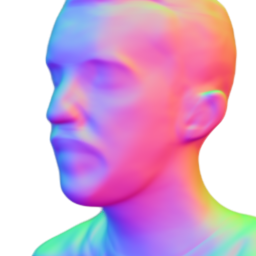}
  \includegraphics[width=0.088\textwidth]{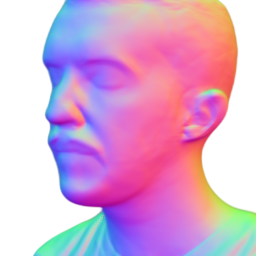}
   \includegraphics[width=0.088\textwidth]{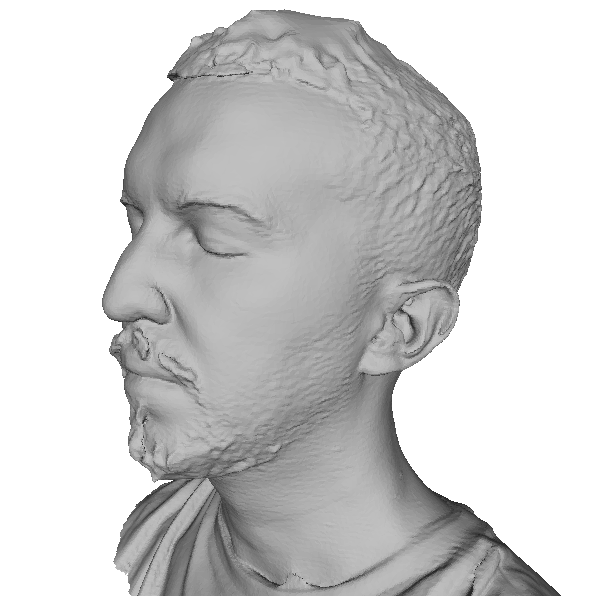}
    \\
    \raisebox{0.4\height}{\makebox[0.01\textwidth]{\rotatebox{90}{\makecell{\scriptsize ID 07}}}}
    \includegraphics[width=0.088\textwidth]{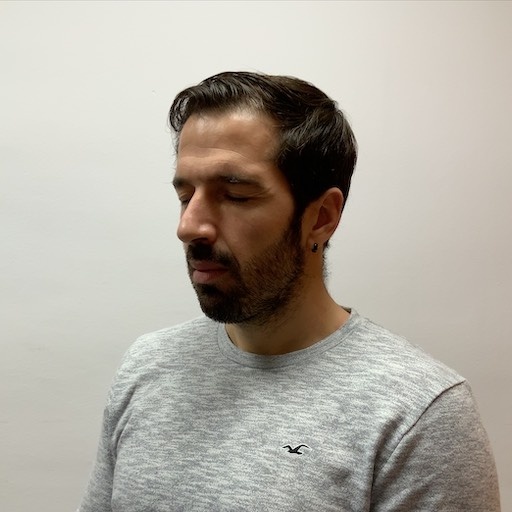}
    \includegraphics[width=0.088\textwidth]{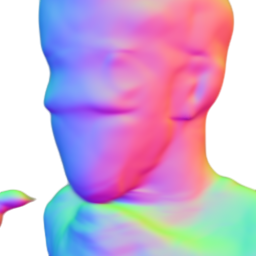}
    \includegraphics[width=0.088\textwidth]{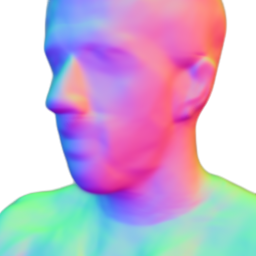}
    \includegraphics[width=0.088\textwidth]{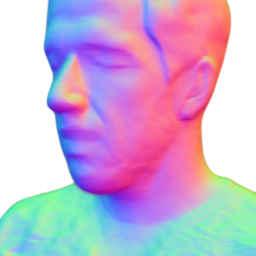}
     \includegraphics[width=0.088\textwidth]{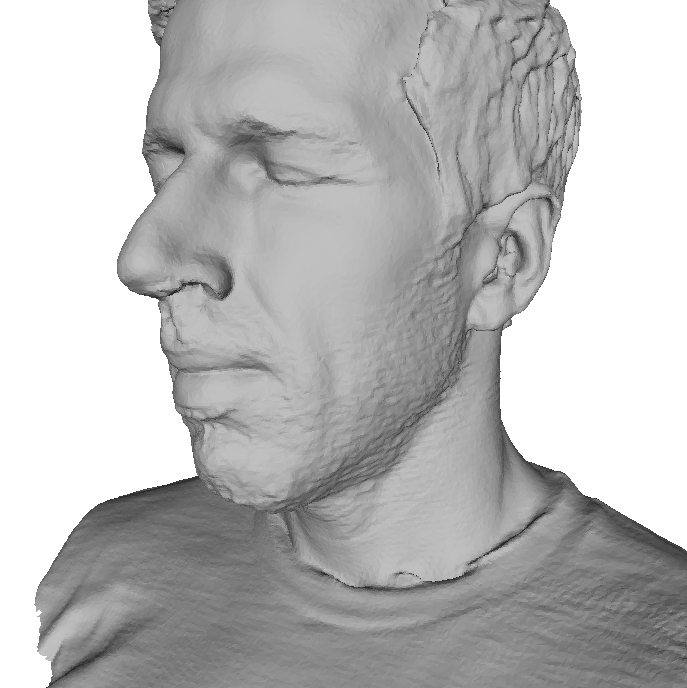}
    \\
    \caption{A two-stage strategy: stage 1 focuses on learning the basic geometry of the head, and stage 2 only uses gradient images to optimize facial geometric details. Even compared with the full RGB image given 30 viewing directions, our reconstruction still achieves comparable results. } 
    \vspace{-0.1in}
    \label{fig:evaluation}
\end{figure}
\vspace{-0.1in}
\paragraph{Baseline.}
Our approach is based on VolSDF~\cite{yariv2021volume} due to its stability in constructing detailed facial geometry. The flexibility of our framework allows for the incorporation of alternative rendering and training methods such as 2DGS~\cite{huang20242d} in Figure~\ref{fig:meeting room}. We evaluate the geometric accuracy of our work under identical camera poses and view directions. VolSDF is based on all full-face images, while our method doesn't rely on sensitive facial images. Given that our method is trained on gradient images, a large amount of information is lost compared to full RGB images at all viewing angles, and our results remain comparable to VolSDF, as detailed in the Evaluation.
\vspace{-0.05in}
\paragraph{Evaluation.}
We evaluate the quality of our algorithm's geometry recovery through both qualitative and quantitative comparisons with the VolSDF algorithm. We present results obtained after the first stage of privacy-neutral training and the second stage of privacy-protected reconstruction, comparing them with VolSDF's full-image supervised reconstruction outcomes in Figure \ref{fig:evaluation}.
To evaluate the overall reconstruction accuracy, we employ the Chamfer distance (CD) metric, computed for both our method and VolSDF. These metrics are detailed in Table \ref{table: table2}. 
Furthermore, our reconstruction excels in capturing intricate details such as hair or interesting facial features such as different expressions in Figure \ref{fig:hair}. This highlights the versatility of our method in handling intricate aspects beyond facial contours.
\newcommand{\CD}{CD\xspace}
\newcommand{\Mean}{Mean\xspace}

\newcommand{\Frst}[1]{\textcolor{red}{\textbf{#1}}}
\newcommand{\Scnd}[1]{\textcolor{blue}{\textbf{#1}}}
\begin{table*} \centering
    \resizebox{\textwidth}{!}{
        \Large
    \begin{tabular}{l||*{1}{c}|*{1}{c}|*{1}{c}|*{1}{c}|*{1}{c}|*{1}{c}|*{1}{c}|*{1}{c}|{c}|*{1}{c}||*{1}{c}}
        \toprule 
         \rowcolor{lightgrey}Method\(\downarrow\)\ & 393 & 421 & 395 & 346  & 340 & 375 & 411 & 393\_4 & 393\_3 & 393\_2 & \textbf{Mean}\\
        \midrule
        Ours stage 1  & $2.7692$ & $1.7467$ & $2.8889$ & $2.3320$ & $1.7957$ & $2.6392$ & $2.2117$ & $2.1790$ & $2.9775$ & $2.0121$ & $2.3552$ \\
        Ours stage 2 & $0.6998$ & $0.7176$ & $0.7645$ & $0.7137$ & $0.8382$ & $0.8289$ & $0.8294$ & $0.8556$ & $0.7798$ & $0.7965$ & $0.7824$ \\
        VolSDF~\cite{yariv2021volume}(full img) & $0.4509$ & $0.5163$ & $0.5399$ & $0.4547$ & $0.4988$ & $0.5266$ & $0.5059$ & $0.4366$ & $0.5394$ & $0.4788$ & $0.4948$ \\
        \bottomrule
    \end{tabular}
    }\\
    \makebox[\linewidth]{\normalsize (a) CD ($10^{-3}$) results on 10 identities in the FaceScape dataset~\cite{yang2020FaceScape}.}
    \resizebox{\textwidth}{!}{
        \Large
    \begin{tabular}{l||*{1}{c}|*{1}{c}|*{1}{c}|*{1}{c}|*{1}{c}|*{1}{c}|*{1}{c}|*{1}{c}|{c}|*{1}{c}||*{1}{c}}
        \toprule
        \rowcolor{lightgrey}Method\(\downarrow\)\ & 00 & 01 & 02 & 03  & 04 & 05 & 06 & 07 & 08 & 09 & \textbf{Mean} \\
        \midrule
        Ours stage 1  & 3.8912 & 3.9796  & 3.508   & 4.1967 & 3.8524  & 3.8205 &  5.9922 
                      & 6.8757 & 4.4531  & 5.0861  & 4.7066  \\
        Ours stage 2      & 3.0397 & 3.1610 & 2.8901 & 3.4117 & 2.9827 & 3.6830 & 3.3911
                      & 3.0951  & 3.1113 & 2.5912 & 3.0477  \\       
        VolSDF~\cite{yariv2021volume} (full imgs)          & 2.7931 & 2.2665 & 2.3108 & 2.5131  & 2.607 & 2.6912  &  3.1767 & 2.7124  & 2.9878  & 2.3938  & 2.6351 \\
        \bottomrule
    \end{tabular}
    }\\
    \vspace{2pt}
    \makebox[\linewidth]{\normalsize (b) CD (mm) results on 10 identities in the H3DS dataset~\cite{ramon2021h3d}.} \vspace{-0.2in}
    \caption{We utilize CD to evaluate the quality of the reconstructed mesh, where lower values indicate better performance. To ensure a fair comparison, we employ the same technique to trim and process the mesh. In comparison to the first stage, our geometric accuracy has improved, and the CD value has been reduced to a level comparable to that of the full RGB image input.
    }
    \label{table: table2}
\end{table*}

\begin{figure}\centering
    \makebox[0.01\textwidth]{}
    \makebox[0.114\textwidth]{\small ground truth}
    \makebox[0.114\textwidth]{\small w/o  Lip.}
    \makebox[0.114\textwidth]{\small w/  Lip.}
    \\
\raisebox{1\height}{\makebox[0.01\textwidth]{\rotatebox{90}{\makecell{\scriptsize ID 393}}}}
    \includegraphics[trim=1cm 2 3cm 1cm, clip,width=0.104\textwidth]{image/evaluation/393gt.png}
    \includegraphics[trim=1cm 2 3cm 1cm, clip,width=0.104\textwidth]{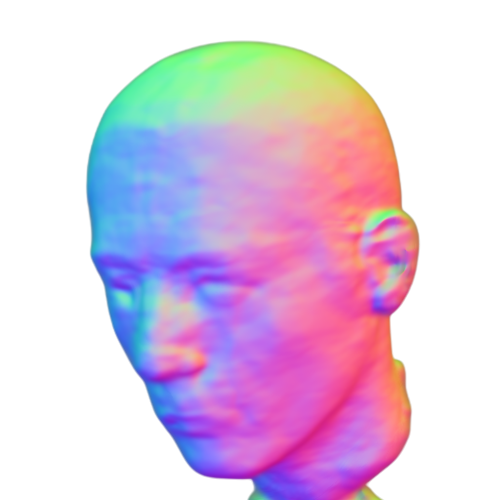}
    \includegraphics[trim=1cm 2 3cm 1cm, clip,width=0.104\textwidth]{image/evaluationnew/393-removebg-preview.png}
    \\
    \raisebox{1\height}{\makebox[0.01\textwidth]{\rotatebox{90}{\makecell{\scriptsize ID 395}}}}
    \includegraphics[trim=1cm 2 3cm 1cm, clip,width=0.104\textwidth]{image/evaluation/395gt.png}
    \includegraphics[trim=1cm 2 3cm 1cm, clip,width=0.104\textwidth]{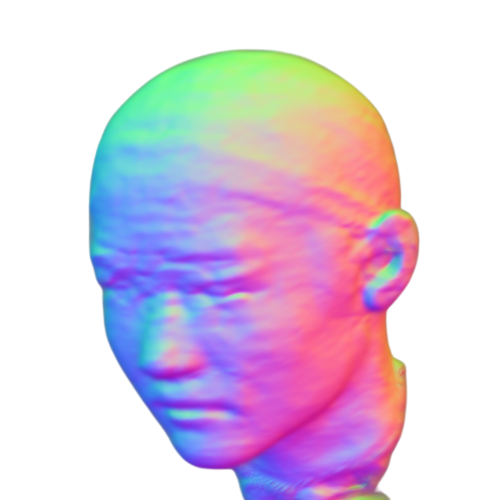}
    \includegraphics[trim=1cm 2 3cm 1cm, clip,width=0.104\textwidth]{image/evaluationnew/395-removebg-preview.png}
    \\
    \vspace{2pt}
    \small
      \resizebox{0.35\textwidth}{!}{
  \begin{tabular}{cc||ccc}
    \toprule
     \rowcolor{lightgrey}& CD ($10^{-3}$)  & ID 393 &ID 395 &  \\
    \midrule
     \rowcolor{lightgreen}& w/ Lip. & 0.6998  & 0.7664 & \\
    & w/o Lip. & 0.7753  & 0.8347 &  \\
    \bottomrule
   \end{tabular}}
    \caption{Lipschitz regularization (Lip.) makes geometry learning more reasonable in Stage 2, such as the mouth (Row 1) and the smoother face (Row 2).} 
    \vspace{-0.1in}
    \label{fig:lip}\vspace{-0.1in}
\end{figure}
\vspace{-0.1in}
\subsection{Ablation studies}
\paragraph{Lipschitz regularization.}
We conduct an ablation study of the Lipschitz regularization as depicted in Figure \ref{fig:lip}. Since the face represents relatively smooth geometry, Lipschitz regularization helps ensure more accurate geometric learning and plays a crucial role in recovering facial details, ultimately resulting in the lowest CD.

\begin{figure}\centering\vspace{-0.1in}
    \makebox[0.01\textwidth]{}
    \makebox[0.2\textwidth]{\small Stage 1}
    \makebox[0.2\textwidth]{\small Stage 2}
    \\
    \raisebox{0.5\height}{\makebox[0.01\textwidth]{\rotatebox{90}{\makecell{\small 15 views}}}}
    \includegraphics[trim=2cm 1cm 3cm 1cm, clip,width=0.093\textwidth]{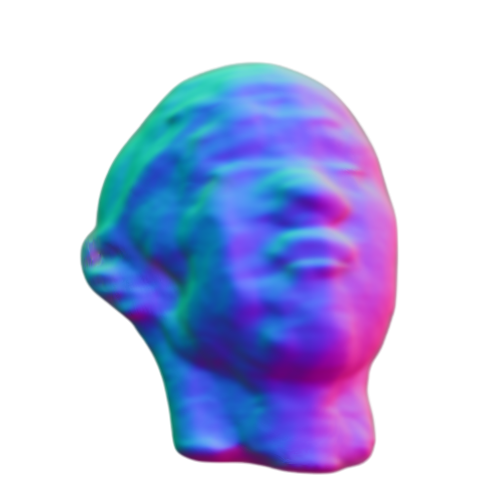}
    \includegraphics[trim=2cm 1cm 3cm 1cm, clip,width=0.093\textwidth]{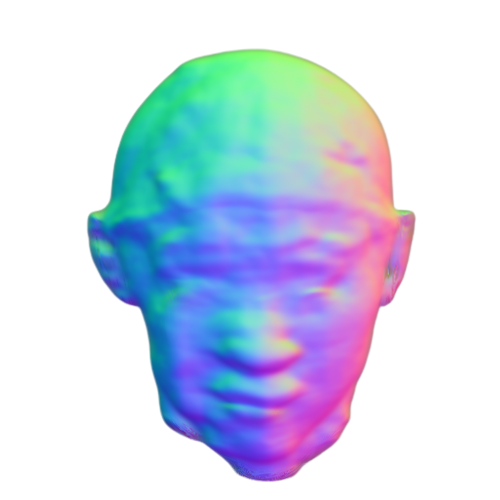}
    \includegraphics[trim=2cm 1cm 3cm 1cm, clip,width=0.093\textwidth]{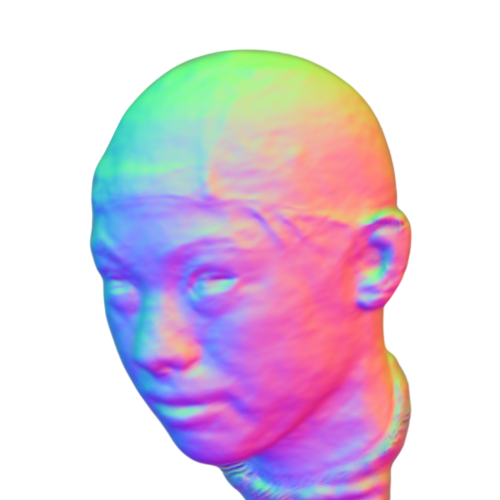}
    \includegraphics[trim=2cm 1cm 3cm 1cm, clip,width=0.093\textwidth]{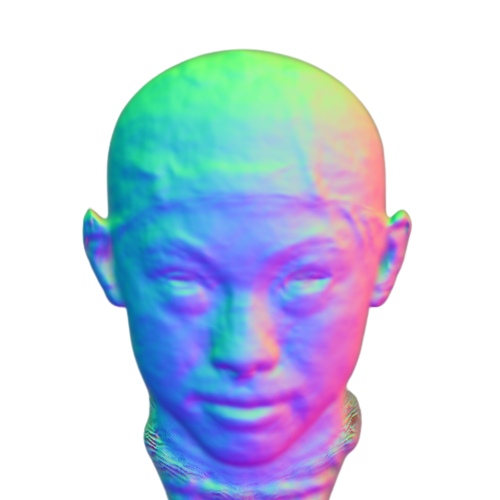}
    \\
    \raisebox{0.5\height}{\makebox[0.01\textwidth]{\rotatebox{90}{\makecell{\small 10 views}}}}
    \includegraphics[trim=2cm 1cm 3cm 1cm, clip,width=0.093\textwidth]{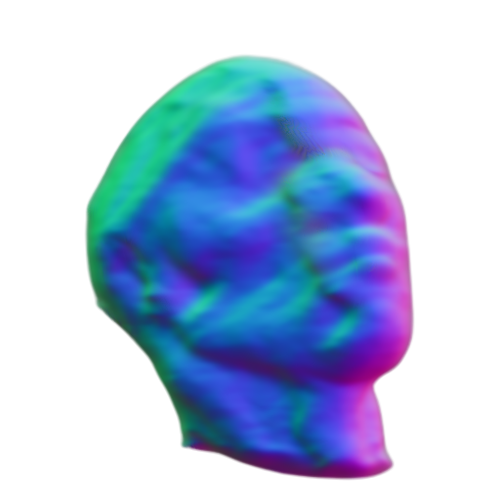}
    \includegraphics[trim=2cm 1cm 3cm 1cm, clip,width=0.093\textwidth]{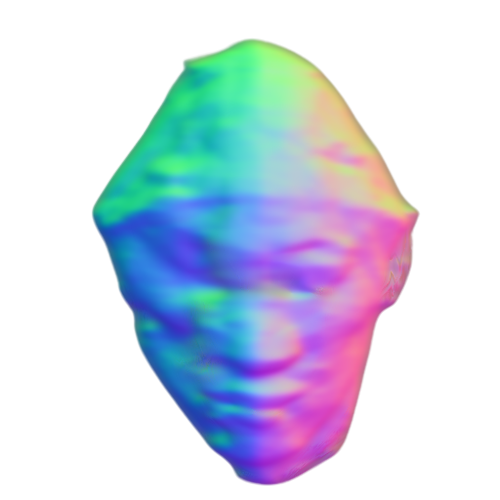}
    \includegraphics[trim=2cm 1cm 3cm 1cm, clip,width=0.093\textwidth]{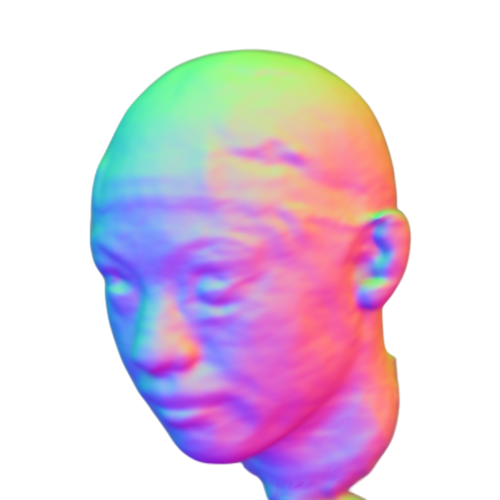}
    \includegraphics[trim=2cm 1cm 3cm 1cm, clip,width=0.093\textwidth]{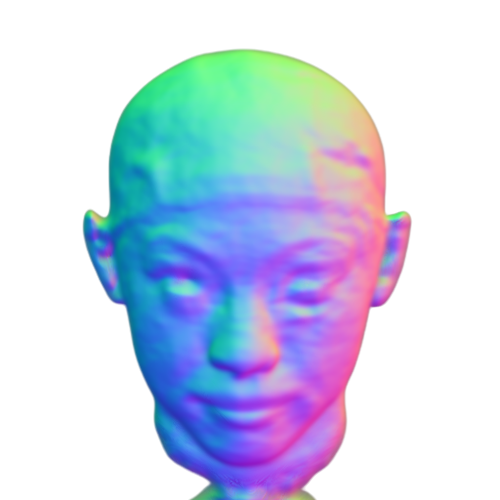}
    \\
    \raisebox{0.5\height}{\makebox[0.01\textwidth]{\rotatebox{90}{\makecell{\small 10 views}}}}
    \includegraphics[trim=2cm 1cm 3cm 1cm, clip,width=0.093\textwidth]{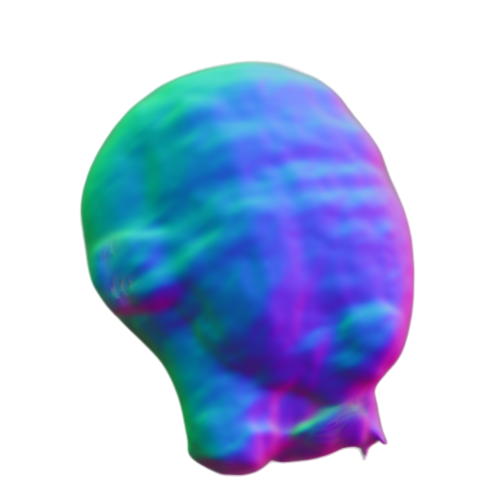}
    \includegraphics[trim=2cm 1cm 3cm 1cm, clip,width=0.093\textwidth]{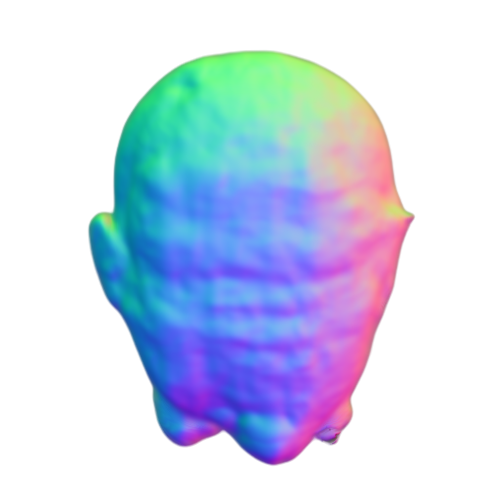}
    \includegraphics[trim=2cm 1cm 3cm 1cm, clip,width=0.093\textwidth]{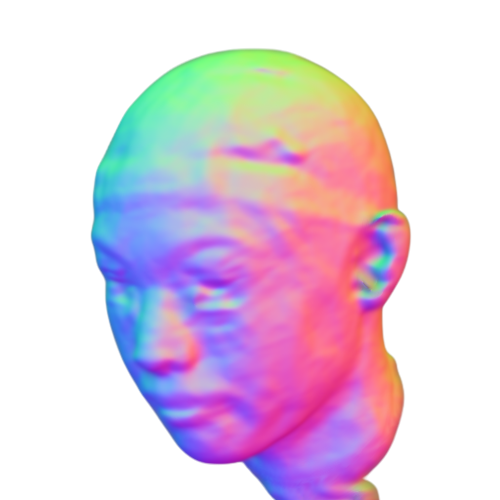}
    \includegraphics[trim=2cm 1cm 3cm 1cm, clip,width=0.093\textwidth]{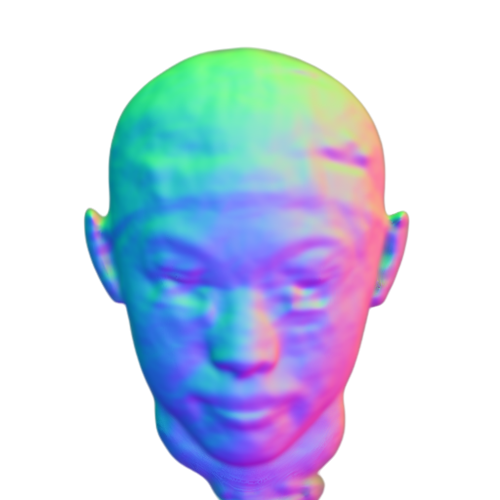}
    \\ 
    \caption{Varying the number of RGB images and views in the first stage may affect Stage 1 accuracy but has minimal impact on final reconstruction quality. Row 1 shows a one-stage process with 15 views, while Rows 2 and 3, both with 10 views, display different input views.}  \vspace{-0.1in}
    \label{fig:view}\vspace{-0.1in}
\end{figure}

\begin{figure} 
\centering\vspace{-0.1in}
    \makebox[0.45\linewidth]{\small Template}
    \makebox[0.45\linewidth]{\small Results}
    \\
    \includegraphics[trim=3cm 0 1cm 0, clip,scale=1.5,width=0.098\textwidth]{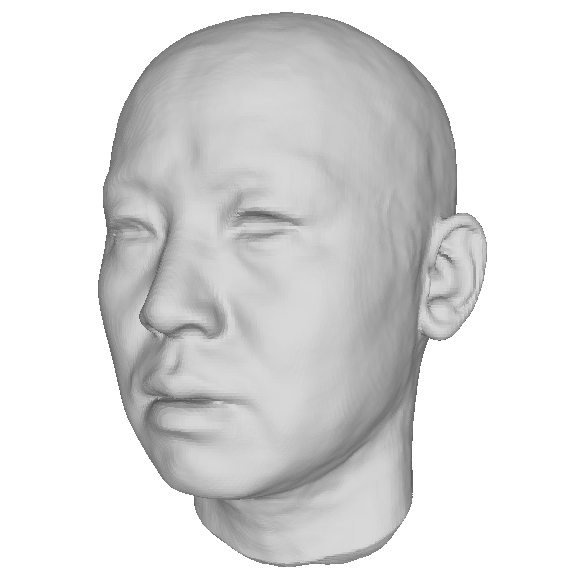}
    \includegraphics[trim=3cm 0 1cm 0, clip,scale=1.5,width=0.098\textwidth]{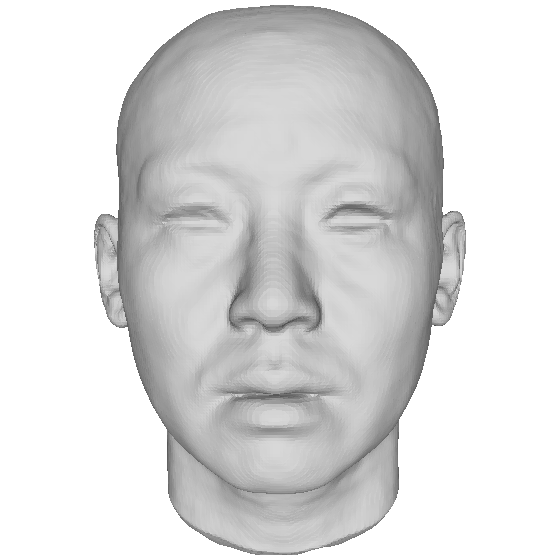}
    \includegraphics[trim=3cm 0 1cm 0, clip,scale=1.5,width=0.098\textwidth]{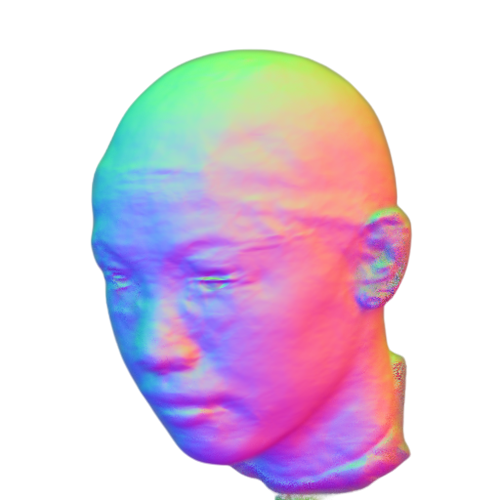}
   \includegraphics[trim=3cm 0 1cm 0, clip,scale=1.5,width=0.098\textwidth]{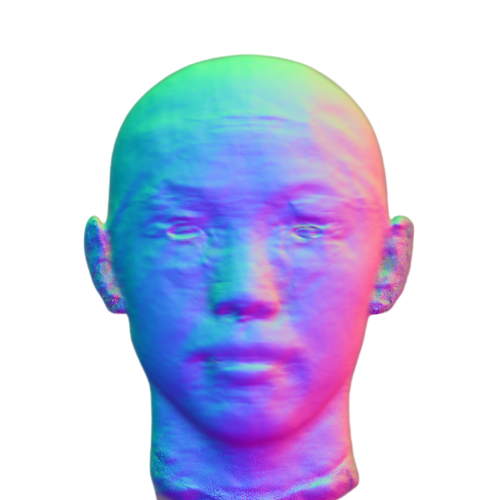}
    \\
    \vspace{-0.1in}\caption{Template. Two on the left showcase the neutral template we employ, while two on the right are our reconstruction results using the template.}  \vspace{-0.1in}
    \label{fig:template}
\end{figure}

\vspace{-0.1in}
\paragraph{Sparse views and geometric priors.}
Considering varying user definitions of privacy, we compare scenarios with fewer color images in Stage 1, as shown in Figure \ref{fig:view}. Stricter privacy requirements may impact Stage 1 accuracy, but the final results remain acceptable. In extreme cases where all provided images contain sensitive features, a one-stage geometric simulation using identity-free templates trained on multiple heads can be used. Figure \ref{fig:template} shows that these templates, while not fully replacing original reconstructions due to missing low-frequency geometric information, are effective in special situations.

\begin{figure}\centering\vspace{-0.1in}
        \includegraphics[width=0.13\textwidth]{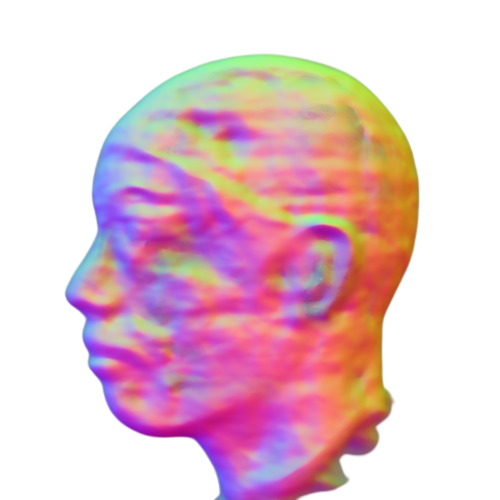}
        \includegraphics[width=0.13\textwidth]{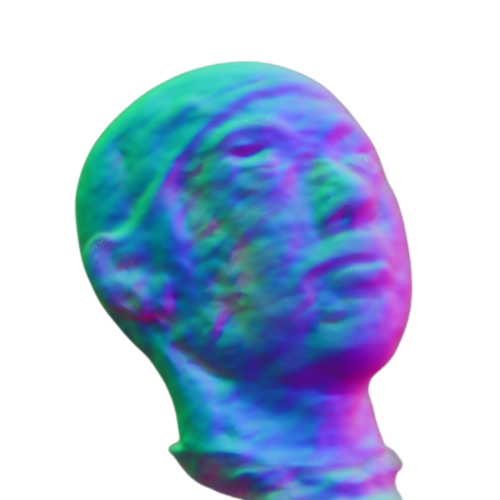}
\includegraphics[width=0.134\textwidth]{image/sparse_view/12-removebg-preview.png}
    \makebox[0.01\textwidth]{}
    \makebox[0.13\textwidth]{\small (a) 27.65}
    \makebox[0.13\textwidth]{\small (b) 15.67}
     \makebox[0.13\textwidth]{\small (c) 0.84}
   \vspace{-0.1in} \caption{Single stage reconstruction with CD ($10^{-3}$) values. (a) shows the single-stage reconstruction using only gradient loss; (b) shows the single-stage training with both rgb and gradient losses; (c) shows our two-stage training result.}
    \vspace{-0.1in}
    \label{fig:fail}\vspace{-0.1in}
\end{figure}

Our experiments show that the first stage is crucial for extracting essential geometric priors, benefiting the second stage significantly. Without this initial stage, reconstructions from gradient examples lack smoothness and plausible geometry, as shown in Figure \ref{fig:fail} (a), due to missing low-frequency information. Training with both privacy-neutral and privacy-protected images on a template, as shown in Figure \ref{fig:fail} (b), yields unsatisfactory results. This occurs because inadequate low-frequency information during the initial training disrupts the learning process when high-frequency data is introduced.
\vspace{-0.1in}

\subsection{Security Discussion}
\begin{figure}\vspace{-0.1in}
  \centering
    \makebox[0.088\textwidth]{\scriptsize Original 11}
    \makebox[0.01\textwidth]{}
    \makebox[0.088\textwidth]{\scriptsize Original 11}
    \makebox[0.088\textwidth]{\scriptsize Original 08}
    \makebox[0.088\textwidth]{\scriptsize Protected 11}
    \makebox[0.088\textwidth]{\scriptsize Mesh 11}
    \\
  \includegraphics[width=0.45\textwidth]{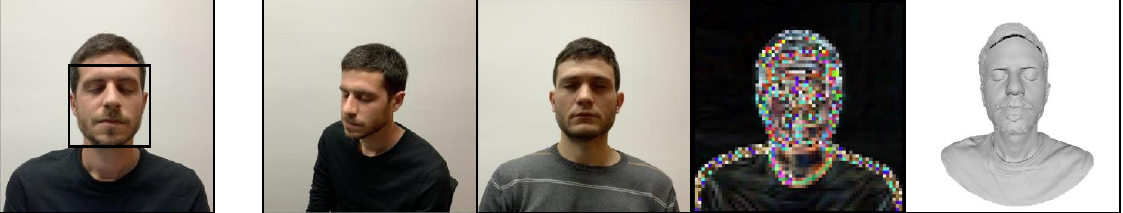}\\ 
   \small
      \resizebox{0.4\textwidth}{!}{
  \begin{tabular}{c||ccccccc}
    \toprule
     Verified & true  & \cellcolor{lightgreen}false & \cellcolor{lightgreen}false & \cellcolor{lightgreen}false\\ 
     Distance & 0.4978
  & \cellcolor{lightgreen}0.6145 &\cellcolor{lightgreen}0.9095&\cellcolor{lightgreen}0.8893  \\
    \bottomrule
   \end{tabular}}
  \vspace{-0.05in}
  \caption{VGG verification of the four images (right) against the original (left), showing cosine distances. Greater distances indicate lower similarity.}\vspace{-0.1in}
  \label{fig:distance_vgg}\vspace{-0.13in}
\end{figure}

\paragraph{Avoiding 2D FR tracking.}
\begin{table}[ht]
    \centering\vspace{-0.1in}
    \resizebox{\halftablewidth}{!}{%
    \begin{tabular}{c||c|c|c|c|c}
        \hline
        \rowcolor{lightgrey} \parbox{1.3cm}{Method} & \parbox{1.3cm}{Original} & \parbox{1.3cm}{Protected}  &  \parbox{1.3cm}{\centering Rendered \\ Mesh} & \parbox{1.3cm}{\centering Our \\ Rendering} & \parbox{1.5cm}{\centering VolSDF \\ Rendering} \\ \hline
        VGG-Face  & 93.75  & 6.51  & 3.03 & 43.74 & 84.62\\ 
        Facenet  & 96.88  & 9.68  &  18.18 & 31.23 & 81.52\\ 
        Facenet512  & 96.87  & 3.23  & 3.20& 18.77& 79.27\\ 
        ArcFace  & 90.63  & 6.66  & 6.46 & 31.25 & 79.23\\ 
        Dlib  & 93.75  & 3.17  & 1.77 & 37.50 & 71.57 \\ 
        SFace  & 84.38  & 6.45  & 9.14 & 43.71& 73.35\\ 
        GhostFaceNet  & 85.42  & 6.62  &  1.04 & 31.44& 73.92 \\ \hline
        Average & 91.67 & \cellcolor{lightgreen}6.05 & \cellcolor{lightgreen}6.12&\cellcolor{lightgreen}33.95&77.64\\  \hline
    \end{tabular}%
    }
    \caption{Comparison of recognition accuracy (\%) across 7 FR systems shows that our protected images, directly rendered mesh images, and our rendered results all significantly reduce recognition accuracy, thus avoiding FR tracking.}
    \label{tab:FR avoid}
\end{table}
Although 3D meshes contain identifiable information, they are more robust against malicious exploitation compared to 2D facial images. This robustness stems from the challenges in collecting large-scale 3D datasets, which has limited the development of deep learning-based 3D face recognition~\cite{guo20233d}. In contrast, most widely deployed FR systems rely on 2D images due to their speed and convenience. The powerful capabilities of 2D FR systems, if misused for tracking and surveillance, can lead to severe privacy concerns. To demonstrate the effectiveness of our privacy-preserving method, we evaluated its resistance against seven state-of-the-art (SOTA) 2D face recognition systems~\cite{parkhi2015deep,schroff2015facenet,deng2019arcface,king2009dlib,zhong2021sface,alansari2023ghostfacenets}, which typically represent faces as vectors, using metrics like cosine similarity (Figure~\ref{fig:distance_vgg}) to measure the similarity between images of the same person.
The tests results in Table~\ref{tab:FR avoid} included original sensitive images (Original), protected images (Protected),  rendering results from both our method (Our Rendering) and VolSDF (VolSDF Rendering), and images directly rendered from 3D geometric meshes (Rendered Mesh). 
which confirms that our approach significantly reduces FR accuracy, demonstrating its effectiveness in preventing tracking and lowering security risks.

\begin{figure}\vspace{-0.1in}
    \centering\vspace{-0.1in}
     \makebox[0.0001\textwidth]{}
        \makebox[0.15\textwidth][c]{\scriptsize GT}
        \makebox[0.15\textwidth][c]{\scriptsize VolSDF}
        \makebox[0.15\textwidth][c]{\scriptsize Ours}
        \makebox[0.0001\textwidth]{}
        \includegraphics[width=0.46\textwidth]{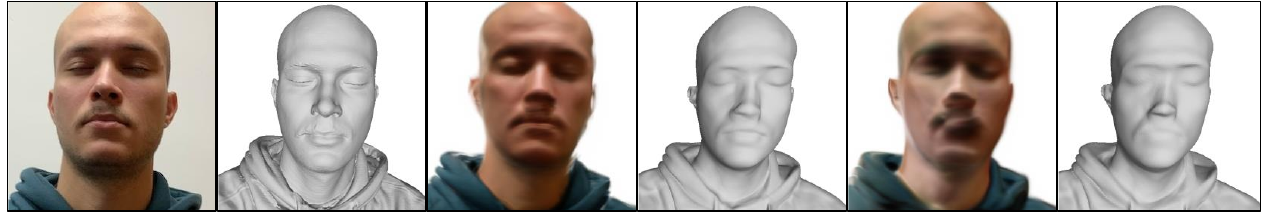}
        \\
    \makebox[0.088\textwidth]{\scriptsize Source}
    \makebox[0.068\textwidth]{\scriptsize Target}
    \makebox[0.088\textwidth]{\scriptsize Origin}
    \makebox[0.088\textwidth]{\scriptsize VolSDF }
    \makebox[0.088\textwidth]{\scriptsize Ours}
    \includegraphics[width=0.45\textwidth]{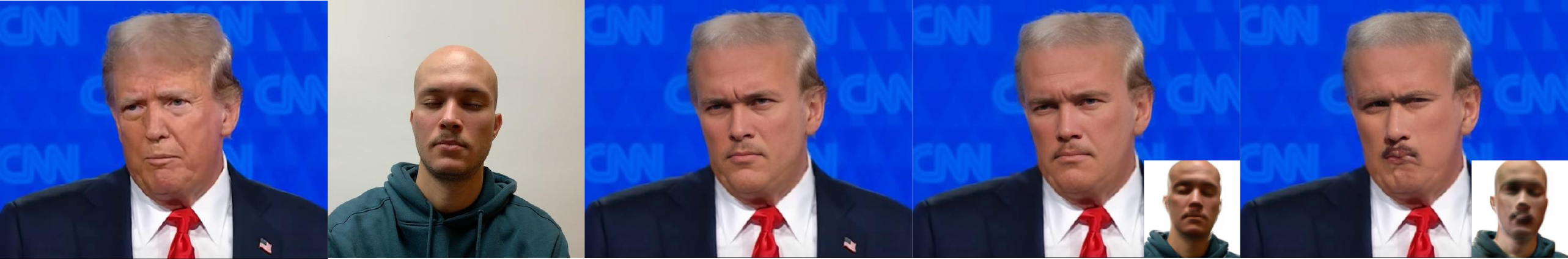}
    \\
    \includegraphics[width=0.45\textwidth]{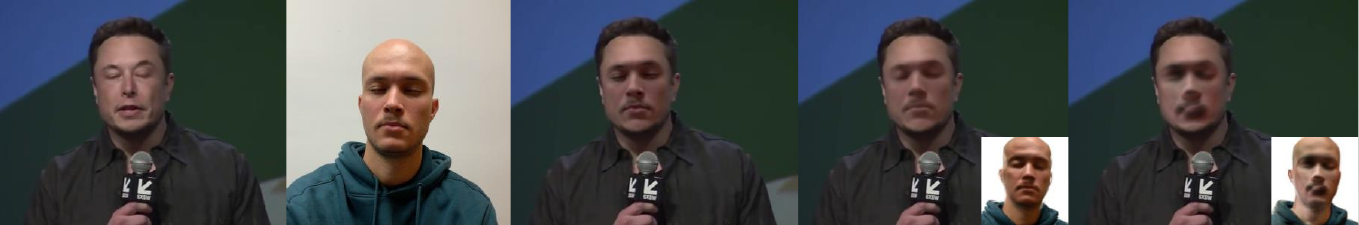}
    \caption{We reconstruct geometric details without rendering realistic sensitive images (row 1). Our rendered images after DeepFake processing retain significantly different facial details from the target person (row 2,3), demonstrating that DeepFake fails to convincingly swap faces with our results thus our method effectively prevents impersonation.}
    \label{fig:deepfake}\vspace{-0.15in}
\end{figure}

\begin{figure}
  \centering\vspace{-0.2in}
  \includegraphics[width=0.45\textwidth]{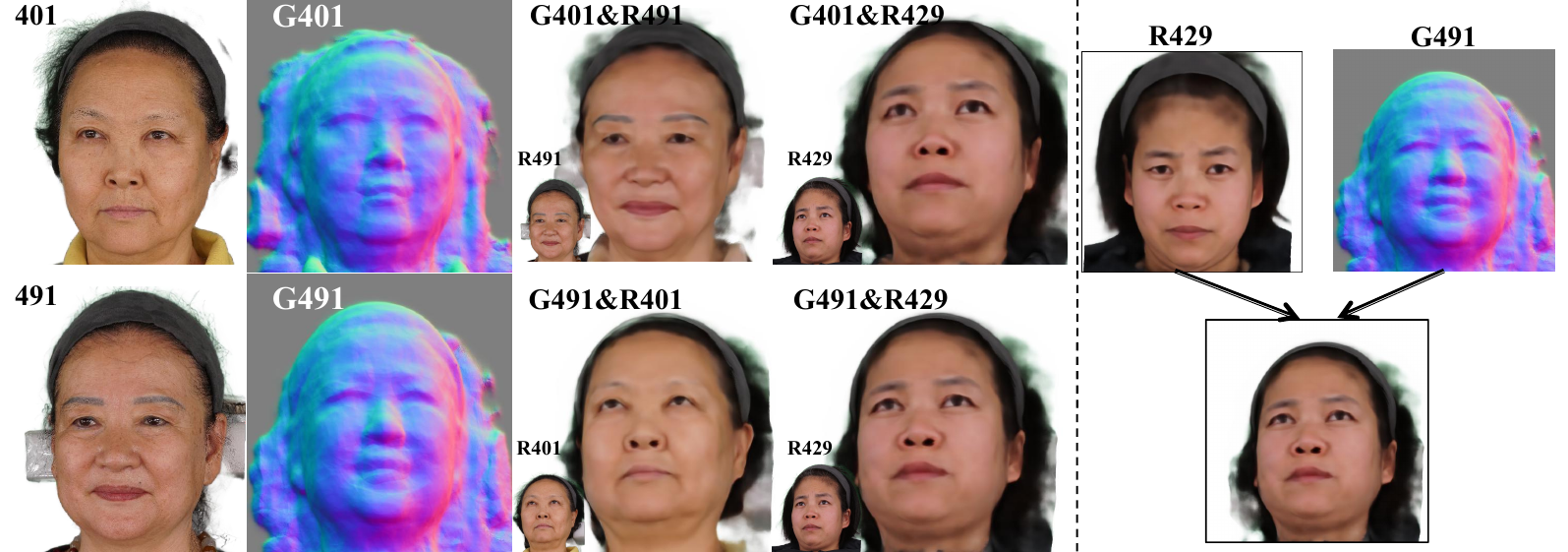}
  \\
  \makebox[0.01\textwidth]{}
    \makebox[0.07\textwidth]{\scriptsize Original }
    \makebox[0.07\textwidth]{\scriptsize Geometry}
    \makebox[0.15\textwidth]{\scriptsize Transfer Results}
    \makebox[0.14\textwidth]{\scriptsize  Transfer    }\vspace{-0.1in}
  \caption{We use $G{\ast}$ to denote the geometry information and use $R{\ast}$ to represent the radiance information of identity~${\ast}$. Each row corresponds to the same geometry with varying radiance applied. Transfer results resemble faces with radiance, with quantitative analysis in the supplementary materials showing that facial texture greatly affects perceived identity.}\vspace{-0.1in}
  \label{fig:texture}
\end{figure}

\paragraph{Counteracting DeepFake.}
Current DeepFake technology can convincingly replace facial details through other people's images and videos, resulting in highly realistic and potentially harmful fabrications. In NeRF-based reconstructions, DeepFake can participate in fabrications in two ways: one requires users to provide multi-view facial images, which our method avoids, and the other applies DeepFake to rendered images. Figure \ref{fig:deepfake} compares results from DeepFake models with (Row 2) and without (Row 3) pre-trained generative models. The inputs include original multi-view RGB images, images rendered with VolSDF, and images rendered with our method. Both types of our DeepFake images fail to convincingly replicate the target individual due to the loss of sensitive facial textures, with quantitative analysis provided in the supplementary materials, making DeepFake technology ineffective.  
\paragraph{Preventing misuse.}
We investigated the role of facial texture in visual recognition by transferring the radiance and geometry from different identities. Figure~\ref{fig:texture} demonstrated that facial mesh transferred with someone else's facial texture radiance appeared significantly more similar to that person, reinforcing the idea that achieving accurate forgery of facial identities requires the acquisition of highly specific, original facial textures. Our findings highlight that without access to the exact texture of the original face, the potential for successful face forgery remains limited.
\paragraph{Future application.}
\begin{figure}
  \centering
  \includegraphics[width=0.42\textwidth]{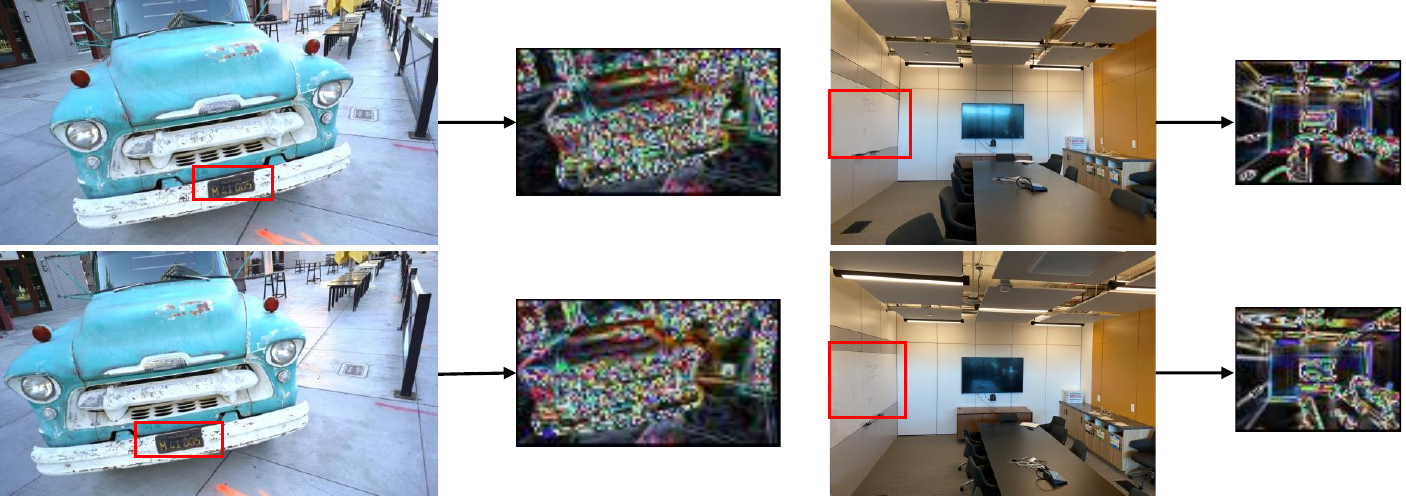}
  \includegraphics[width=0.4\textwidth]{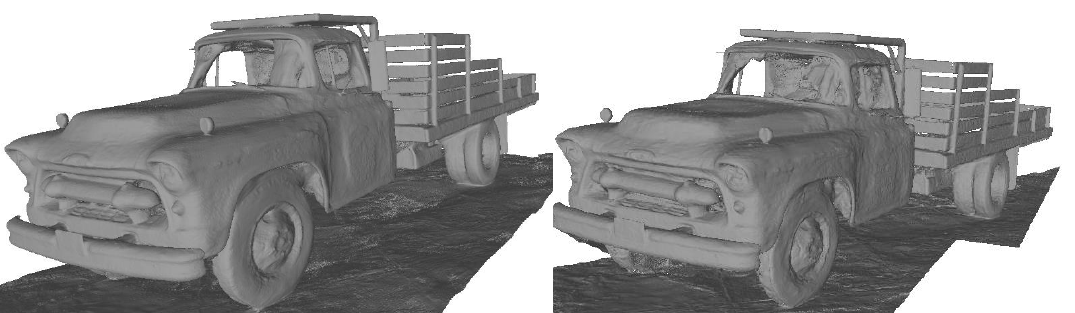}
  \makebox[0.01\textwidth]{}
    \makebox[0.23\textwidth]{\scriptsize Full images reconstruction (2DGS) }
    \makebox[0.23\textwidth]{\scriptsize Protected images reconstruction (2DGS)}
  \caption{For sensitive images (license plates, conference room details), users can provide only protected inputs instead of the actual images, and reconstruct details.}\vspace{-0.1in}
  \label{fig:meeting room}\vspace{-0.1in}
\end{figure}
For 3D reconstruction tasks, our proposed method prevents reliance on sensitive images by achieving high-fidelity geometric reconstruction using only sensitive-image gradients as inputs. This finding is applicable to other neural reconstruction pipelines \cite{huang20242d} and extends beyond facial reconstruction to various scenarios requiring minimal input such as meeting rooms or scenes involving protected elements like vehicle license plates (Figure~\ref{fig:meeting room}). Notably, even with such minimal input, the quality of geometric reconstruction remains comparable.

\vspace{-0.1in}
\section{Conclusion}
In this paper, we highlight the often-overlooked aspect of privacy in neural facial reconstruction. We present a method that reconstructs detailed head geometry without relying on sensitive input. By processing sensitive facial data and focusing on essential geometric information in a two-stage process, our approach balances privacy protection with reconstruction quality. Our method represents a significant advancement, integrating privacy and neural facial reconstruction and paving the way for new explorations in this field.

\section*{Appendix}
\section{Privacy Quantitative experiments}
We have provided visual results in the paper and will discuss the findings in detail, along with presenting quantitative results to further demonstrate the effectiveness of our privacy-protection approach.  
\subsection{DeepFake}
Deep learning has revolutionized computer vision, enabling highly realistic digital image manipulation. A prominent example is face swapping, where a source face is seamlessly transferred onto a target, retaining the target's facial movements and expressions. DeepFake technology, now highly advanced, can achieve film-quality effects, with DeepFakeLab~\cite{perov2020deepfacelab} being the leading software for creating such content. In our study, we first trained the model for 80k iterations without the assistance of a generative model to swap Elon Musk's face. Second, with the rise of generative models, many DeepFake techniques now utilize source expressions and attributes while applying new facial features to make the generated face closely resemble the target individual, so we also conducted face swaps using a generative model. In our paper, we present two sets of results, comparing cases with and without the generative model. Visually, the high realism of neural rendering allows both VolSDF-rendered outputs and the original input images to be clearly identified as the target identity in DeepFakes. However, our rendered results do not contain sensitive facial features, and the DeepFakes generated from them lack clear identity-specific traits.

Face recognition systems~\cite{parkhi2015deep,schroff2015facenet,amos2016openface,king2009dlib,zhong2021sface,alansari2023ghostfacenets} can measure similarity using cosine distance, where a smaller value indicates a higher similarity between two images. For specific implementation details, we referred to~\cite{serengil2020lightface}. For more results and quantitative analysis, please refer to Table~\ref{tab:FR_combined}. DeepFake videos can be found in the supplementary video we provide.
\begin{table*}[ht]
    \centering   
    \begin{minipage}[t]{\textwidth}
    \begin{subtable}[t]{0.48\textwidth}
        \centering
    \centering
    \makebox[0.18\textwidth]{\scriptsize Source}
    \makebox[0.13\textwidth]{\scriptsize Target}
    \makebox[0.18\textwidth]{\scriptsize Origin}
    \makebox[0.18\textwidth]{\scriptsize VolSDF }
    \makebox[0.18\textwidth]{\scriptsize Ours}
    \\
    \includegraphics[width=0.96\textwidth]{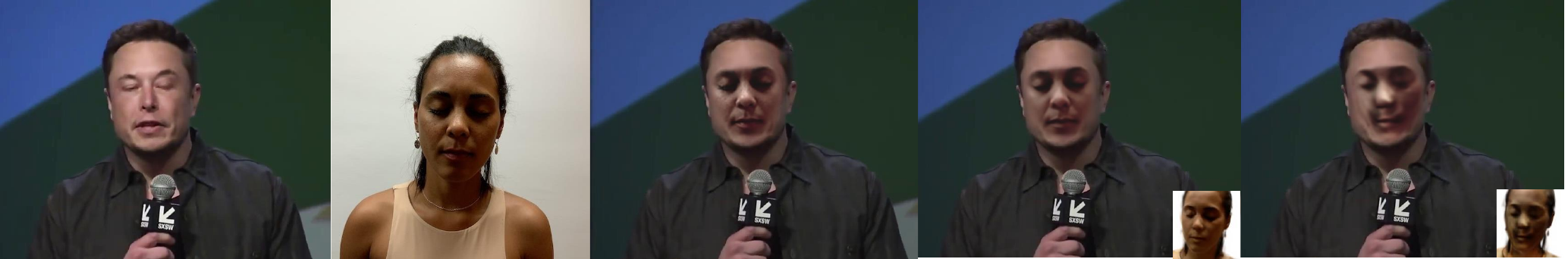}
    \\
    \resizebox{\textwidth}{!}{%
    \begin{tabular}{c||c|c|c}
        \hline
        \rowcolor{lightgrey} \parbox{1.3cm}{Method} & \parbox{1.3cm}{Original} & \parbox{1.3cm}{VolSDF}   & \parbox{1.3cm}{\centering Our }   \\ \hline
        VGG-Face  & 0.5697& 0.5758 & \cellcolor{lightgreen}0.7431 \\ 
        Facenet  & 0.6132  &0.5794  &  \cellcolor{lightgreen}0.659\\ 
        Facenet512  & 0.4068  & 0.4345  &\cellcolor{lightgreen}0.5973 \\ 
        OpenFace &  0.3548   & 0.3566   & \cellcolor{lightgreen}0.4112      \\
        ArcFace  & 0.3523  & 0.3918  & \cellcolor{lightgreen}0.9121 \\ 
        Dlib  & 0.0413  & 0.0391  & \cellcolor{lightgreen}0.1153  \\ 
        SFace  & 0.7191  & 0.7154  & \cellcolor{lightgreen}0.7513 \\ 
        GhostFaceNet  & 0.6288  & 0.6618  & \cellcolor{lightgreen} 0.9541  \\ \hline
    \end{tabular}%
    }   
    \caption{DeepFake without generation model.}
    \label{tab:FR avoid1}
\end{subtable}
\hfill
    \begin{subtable}[t]{0.48\textwidth}
    \centering
    \makebox[0.18\textwidth]{\scriptsize Source}
    \makebox[0.13\textwidth]{\scriptsize Target}
    \makebox[0.18\textwidth]{\scriptsize Origin}
    \makebox[0.18\textwidth]{\scriptsize VolSDF }
    \makebox[0.18\textwidth]{\scriptsize Ours}
    \\
    \includegraphics[width=0.96\textwidth]{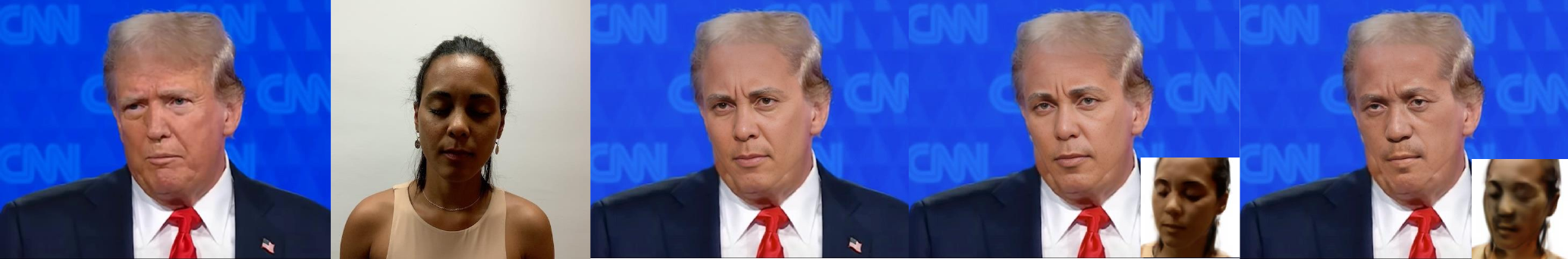}
    \\
    \resizebox{\textwidth}{!}{%
    \begin{tabular}{c||c|c|c}
        \hline
        \rowcolor{lightgrey} \parbox{1.3cm}{Method} & \parbox{1.3cm}{Original} & \parbox{1.3cm}{VolSDF}   & \parbox{1.3cm}{\centering Our }   \\ \hline
         VGG-Face  & 0.5881&0.6133& \cellcolor{lightgreen}0.668 \\ 
        Facenet  & \cellcolor{lightgreen}0.6601  &0.6365 &  0.4443\\ 
        Facenet512  & 0.3294 & 0.3351  &\cellcolor{lightgreen}0.4744 \\ 
        OpenFace &  0.3732   & 0.3482   & \cellcolor{lightgreen}0.4237      \\
        ArcFace   & 0.3345& 0.3166 & \cellcolor{lightgreen}0.4542 \\ 
       Dlib  & 0.0478  &0.0422  &  \cellcolor{lightgreen}0.0661\\ 
        SFace  & 0.5457  & 0.4976  &\cellcolor{lightgreen}0.5953 \\ 
        GhostFaceNet &  0.3538   & 0.3665   & \cellcolor{lightgreen}0.5030      \\ \hline 
    \end{tabular}%
    }
    \caption{DeepFake with generation model. }
    \label{tab:FR avoid4}
\end{subtable}
\end{minipage}
\begin{minipage}[t]{\textwidth}
\begin{subtable}[t]{0.48\textwidth}
    \centering\vspace{0.1in}
    \makebox[0.18\textwidth]{\scriptsize Source}
    \makebox[0.13\textwidth]{\scriptsize Target}
    \makebox[0.18\textwidth]{\scriptsize Origin}
    \makebox[0.18\textwidth]{\scriptsize VolSDF }
    \makebox[0.18\textwidth]{\scriptsize Ours}
    \\
    \includegraphics[width=0.96\textwidth]{image/musk.pdf}
    \\
    \resizebox{\textwidth}{!}{%
    \begin{tabular}{c||c|c|c}
        \hline
        \rowcolor{lightgrey} \parbox{1.3cm}{Method} & \parbox{1.3cm}{Original} & \parbox{1.3cm}{VolSDF}   & \parbox{1.3cm}{\centering Our }   \\ \hline
        VGG-Face  & 0.468& 0.6018 &\cellcolor{lightgreen} 0.6027 \\ 
        Facenet  & 0.2067  &0.3735  & \cellcolor{lightgreen} 0.5196\\ 
        Facenet512  & 0.1376  & 0.3197  &\cellcolor{lightgreen}0.4668 \\ 
        OpenFace &  0.1632   & 0.2264   & \cellcolor{lightgreen}0.3766      \\
        ArcFace  & 0.3247  & 0.5336  &\cellcolor{lightgreen} 0.667 \\ 
        Dlib  & 0.0502  & \cellcolor{lightgreen}0.0781  & 0.0695  \\ 
        SFace  & 0.5342  & \cellcolor{lightgreen}0.6945  & 0.6786 \\ 
        GhostFaceNet  & 0.3895  & 0.5549  & \cellcolor{lightgreen} 0.6571  \\ \hline
    \end{tabular}%
    }   
    \caption{DeepFake without generation model.}
    \label{tab:FR avoid2}
\end{subtable}
\hfill
\begin{subtable}[t]{0.48\textwidth}
    \centering\vspace{0.1in}
    \makebox[0.18\textwidth]{\scriptsize Source}
    \makebox[0.13\textwidth]{\scriptsize Target}
    \makebox[0.18\textwidth]{\scriptsize Origin}
    \makebox[0.18\textwidth]{\scriptsize VolSDF }
    \makebox[0.18\textwidth]{\scriptsize Ours}
    \\
    \includegraphics[width=0.96\textwidth]{image/deepfake_blue.pdf}
    \\
    \resizebox{\textwidth}{!}{%
    \begin{tabular}{c||c|c|c}
        \hline
        \rowcolor{lightgrey} \parbox{1.3cm}{Method} & \parbox{1.3cm}{Original} & \parbox{1.3cm}{VolSDF}   & \parbox{1.3cm}{\centering Our }   \\ \hline
        VGG-Face  & 0.6913& 0.7852 & \cellcolor{lightgreen}0.8501 \\ 
        Facenet  & 0.359  &\cellcolor{lightgreen}0.5103  &  0.4897\\ 
        Facenet512  & 0.3467  & 0.5215  &\cellcolor{lightgreen}0.536 \\ 
        OpenFace &  0.1966   & 0.3026   & \cellcolor{lightgreen}0.352      \\
        ArcFace  & 0.4838  & 0.5907  & \cellcolor{lightgreen}0.6794 \\ 
        Dlib  & 0.0435  & 0.051  & \cellcolor{lightgreen}0.0867  \\ 
        SFace  & 0.6056  & 0.6127  & \cellcolor{lightgreen}0.7718 \\ 
        GhostFaceNet  & 0.4732  & 0.6102  & \cellcolor{lightgreen} 0.7356  \\ \hline
    \end{tabular}%
    }   
    \caption{DeepFake with generation model.}
    \label{tab:FR avoid3}
\end{subtable}
 \end{minipage}
\caption{Comparison of the original image, original image VolSDF rendering, and our rendering after DeepFake, with both quantitative and qualitative analysis. A higher value indicates a lower similarity to the target. Our method often achieves the lowest similarity, effectively preventing DeepFake.}
    \label{tab:FR_combined}
\end{table*}

\subsection{Appearance Transfer}

\noindent\textbf{Original State:}
\begin{align*}
&\text{Person A has facial geometry } G_A \text{ and radiance information } R_A. \\
&\text{Person B has facial geometry } G_B \text{ and radiance information } R_B.
\end{align*}

\noindent\textbf{After Appearance Transfer:}
\begin{align*}
&A' = \text{Transfer}(G_A, R_B) \\
&B' = \text{Transfer}(G_B, R_A)
\end{align*}

\noindent\textbf{Visual Changes:}
\begin{align*}
&\text{ } A' \text{ now visually appears more like Person B.} \\
&\text{} B' \text{ now visually appears more like Person A.}
\end{align*}
We employ the method proposed by~\cite{Xu_2023_ICCV} for achieving appearance transfer. This approach is straightforward and does not require additional operations; it simply aligns the radiance and geometry information in three dimensions without undergoing any additional retraining, ensuring the fairness of this alignment. 

\begin{figure}\centering
        \centering    
         \makebox[0.01\textwidth]{}
    \makebox[0.135\textwidth]{\small Transferred Image }
    \makebox[0.135\textwidth]{\small Geometry provider}
    \makebox[0.135\textwidth]{\small Radiance provider}
    \\
    \vspace{0.1in}
         \makebox[0.01\textwidth]{}
    \makebox[0.135\textwidth]{\small G401\text{\&}R429 }
    \makebox[0.132\textwidth]{\small 401}
    \makebox[0.135\textwidth]{\small 429}
        \\
        \includegraphics[width=0.135\textwidth]{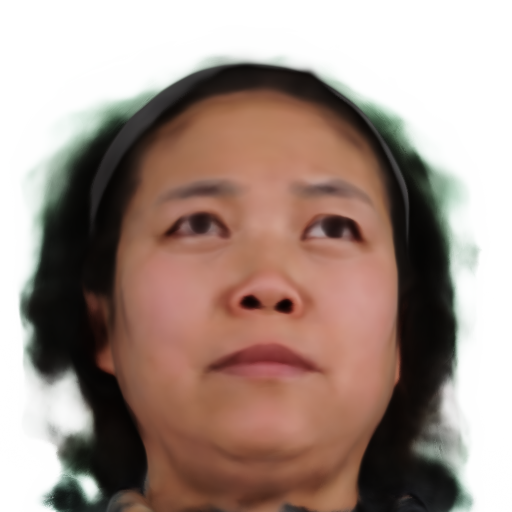}
        \includegraphics[width=0.135\textwidth]{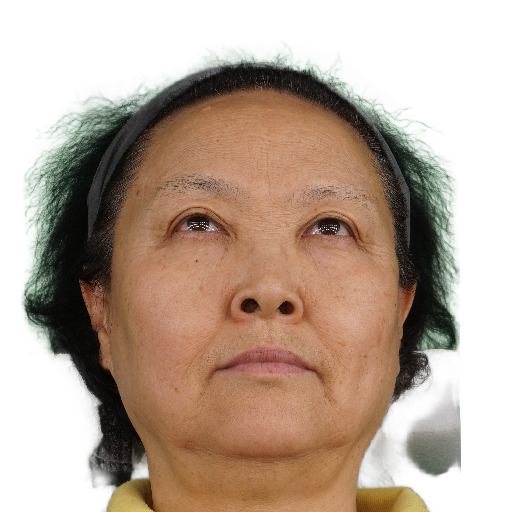}
        \includegraphics[width=0.135\textwidth]{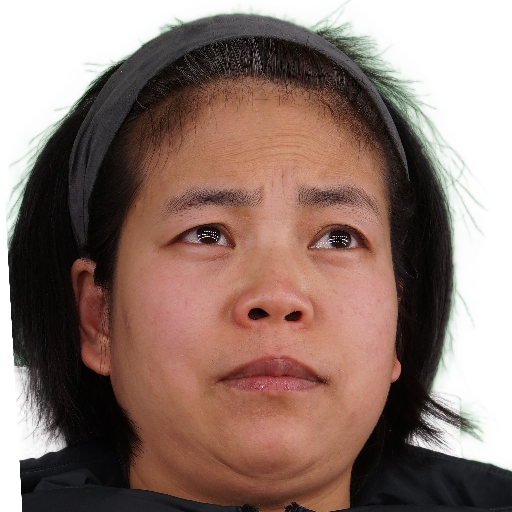}
                 \\
    \makebox[0.01\textwidth]{}
    \makebox[0.135\textwidth]{\small  }
    \makebox[0.135\textwidth]{\small \color{red}{False / 0.6360}}
    \makebox[0.135\textwidth]{\small \color{blue}{True / 0.2570}}
     \makebox[0.01\textwidth]{}
    \makebox[0.135\textwidth]{\small G401\text{\&}R491 }
    \makebox[0.132\textwidth]{\small 401}
    \makebox[0.135\textwidth]{\small 491}
        \\
        \includegraphics[width=0.135\textwidth]{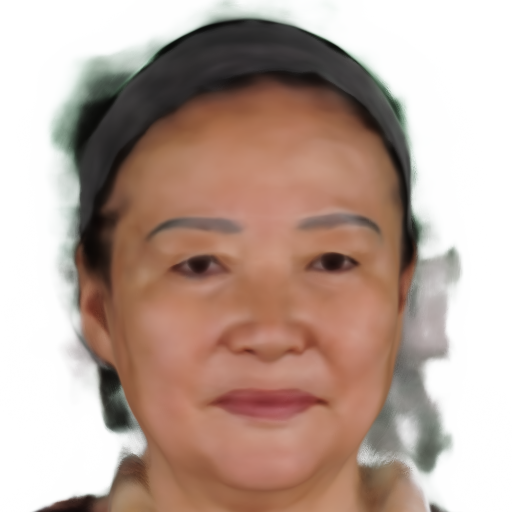}
        \includegraphics[width=0.135\textwidth]{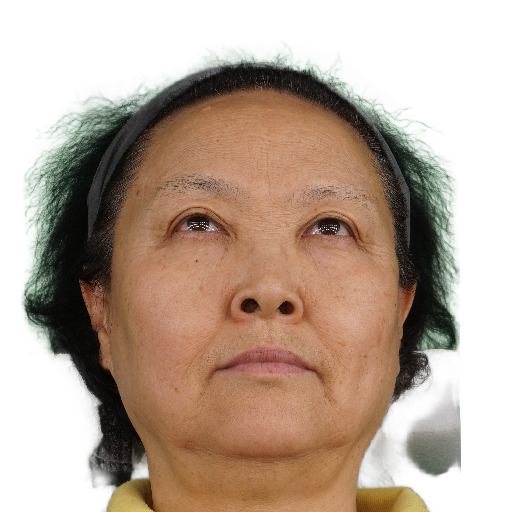}
        \includegraphics[width=0.135\textwidth]{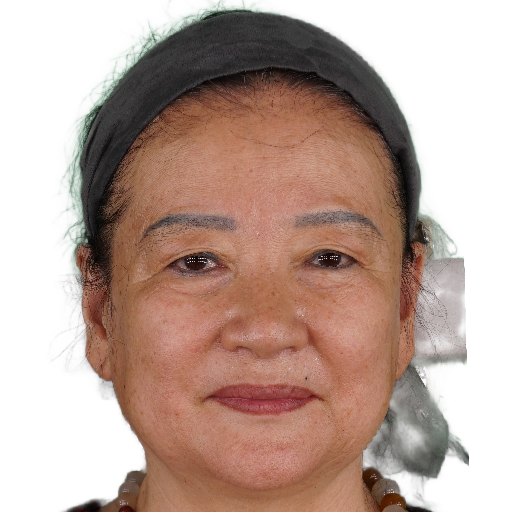}
                 \\
    \makebox[0.01\textwidth]{}
    \makebox[0.135\textwidth]{\small  }
    \makebox[0.135\textwidth]{\small \color{red}{False / 0.4295}}
    \makebox[0.135\textwidth]{\small \color{blue}{True / 0.1109}}
     \makebox[0.01\textwidth]{}
    \makebox[0.135\textwidth]{\small G429\text{\&}R401 }
    \makebox[0.132\textwidth]{\small 429}
    \makebox[0.135\textwidth]{\small 401}
        \\
        \includegraphics[width=0.135\textwidth]{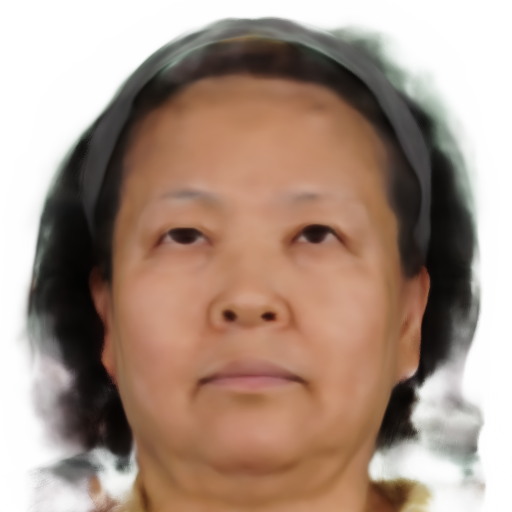}
        \includegraphics[width=0.135\textwidth]{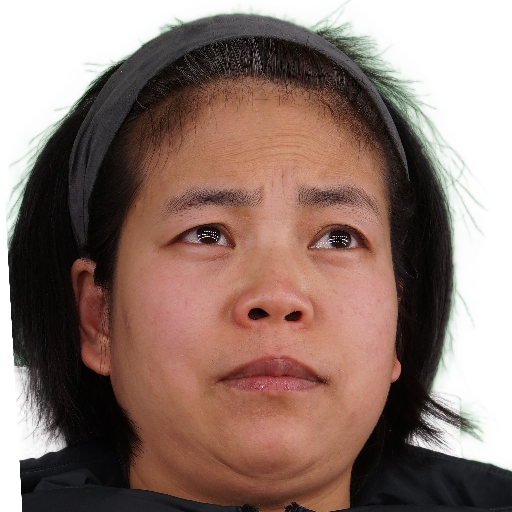}
        \includegraphics[width=0.135\textwidth]{image/0tex.png}
                 \\
    \makebox[0.01\textwidth]{}
    \makebox[0.135\textwidth]{\small  }
    \makebox[0.135\textwidth]{\small \color{red}{False / 0.4728}}
    \makebox[0.135\textwidth]{\small \color{blue}{True / 0.2308}}
     \makebox[0.01\textwidth]{}
    \makebox[0.135\textwidth]{\small G429\text{\&}R491 }
    \makebox[0.132\textwidth]{\small 429}
    \makebox[0.135\textwidth]{\small 491}
        \\
         \includegraphics[width=0.135\textwidth]{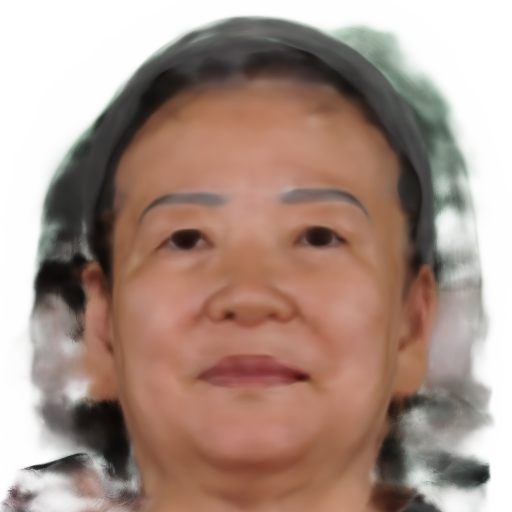}
        \includegraphics[width=0.135\textwidth]{image/1tex.png}
        \includegraphics[width=0.135\textwidth]{image/2tex.png}
                 \\
    \makebox[0.01\textwidth]{}
    \makebox[0.135\textwidth]{\small }
    \makebox[0.135\textwidth]{\small \color{red}{False / 0.5468}}
    \makebox[0.135\textwidth]{\small \color{blue}{True / 0.2183}}
    \hspace{2pt}
     \makebox[0.01\textwidth]{}
    \makebox[0.135\textwidth]{\small G491\text{\&}R429 }
    \makebox[0.132\textwidth]{\small 491}
    \makebox[0.135\textwidth]{\small 429}
        \\
        \includegraphics[width=0.135\textwidth]{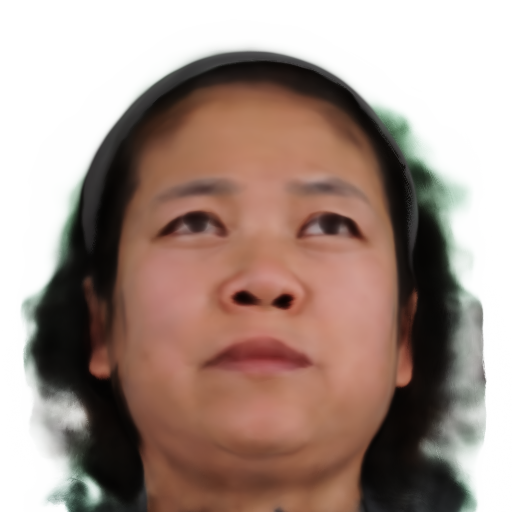}
        \includegraphics[width=0.135\textwidth]{image/2tex.png}
        \includegraphics[width=0.135\textwidth]{image/1tex.png}
                 \\
    \makebox[0.01\textwidth]{}
    \makebox[0.135\textwidth]{\small  }
    \makebox[0.135\textwidth]{\small\color{red}{False / 0.5964}}
    \makebox[0.135\textwidth]{\small \color{blue}{True / 0.1534}}
     \makebox[0.01\textwidth]{}
     \hspace{3pt}
    \makebox[0.135\textwidth]{\small G491\text{\&}R401 }
    \makebox[0.132\textwidth]{\small 491}
    \makebox[0.135\textwidth]{\small 401}
        \\
         \includegraphics[width=0.135\textwidth]{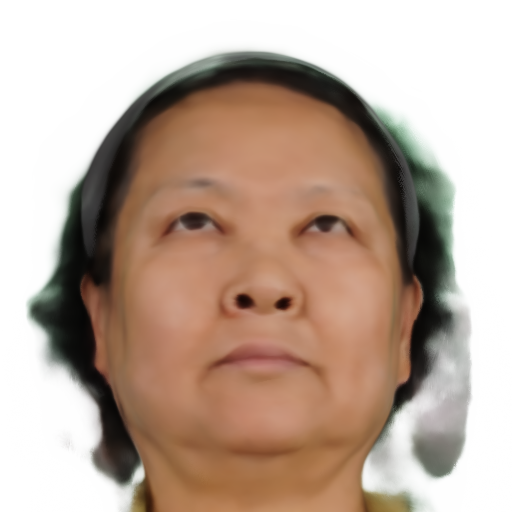}
        \includegraphics[width=0.135\textwidth]{image/2tex.png}
        \includegraphics[width=0.135\textwidth]{image/0tex.png}
                 \\
    \makebox[0.01\textwidth]{}
    \makebox[0.135\textwidth]{\small  }
    \makebox[0.135\textwidth]{\small \color{red}{False / 0.4804}}
    \makebox[0.135\textwidth]{\small \color{blue}{True / 0.2512}}
    
    \caption{The values represent the distances to the transferred image, a smaller distance means more similarity. After transferring, the face is recognized as the identity of the skin texture rather than its original geometry provider, as evaluated using Facenet~\cite{schroff2015facenet}.}
    \label{facenet-PR}
\end{figure}

In the Preventing Misuse section, we discussed that without accurate sensitive facial features, even obtaining the geometry alone does not enable forgery. Here, we provide specific cosine distance values in Figure~\ref{facenet-PR} to demonstrate that, in addition to visual similarity, the transferred results are numerically closer to the radiance results. This further confirms that our proposed method effectively prevents misuse by avoiding the inclusion of sensitive facial features.
\section{Protecting Operators}
\subsection{Privacy-protected argumentations}\label{3.3}
\begin{figure}
  \centering
  \includegraphics[width=0.35\textwidth]{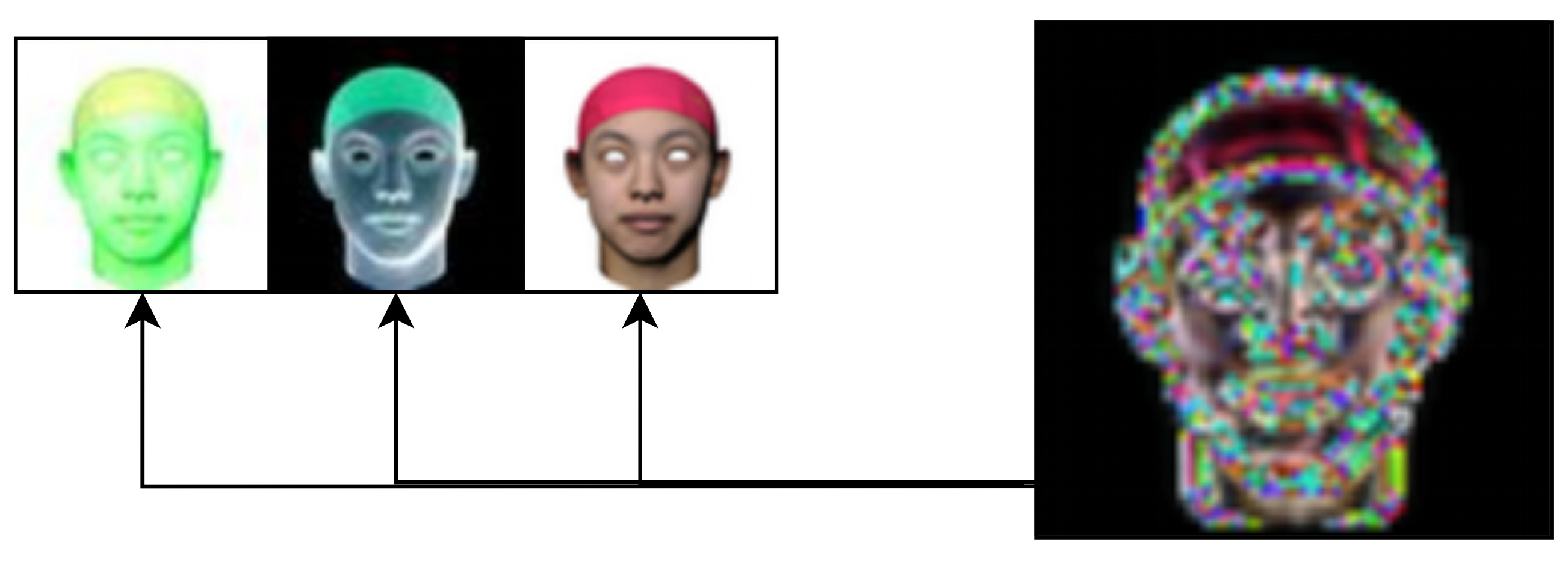}
  \caption{The blurry color gradient modulus derived from the low-resolution GT image can correspond to numerous other images, including those on the left, and not solely the GT image. }   
  \label{fig:multi-mapping} 

\end{figure}

The purpose of the preserving operator is to truncate the user's private information directly on the user's end, without exposing it to anyone. This process requires both simplicity and convenience while ensuring the highest level of privacy protection. To safeguard user privacy, we analyze three key aspects using the preserving operator: 
\paragraph{Irreversibility.} To safeguard privacy, we retain only color variation and discard complete RGB data from color images. This privacy-conscious approach preserves essential data for reconstruction. During image uploads, we retain only the gradient magnitude information $\left\| \mathbf{g} \right\|$, making it practically impossible to reverse engineer RGB data due to many-to-one mapping relationships. 
\begin{figure} \centering
    \makebox[0.01\textwidth]{}
    \makebox[0.144\textwidth]{\small (a)}
    \makebox[0.144\textwidth]{\small (b)}
    \makebox[0.136\textwidth]{\small (c)}
    \\
    \includegraphics[width=0.144\textwidth]{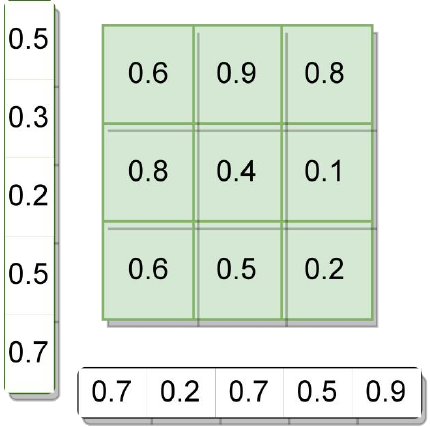}
    \includegraphics[width=0.144\textwidth]{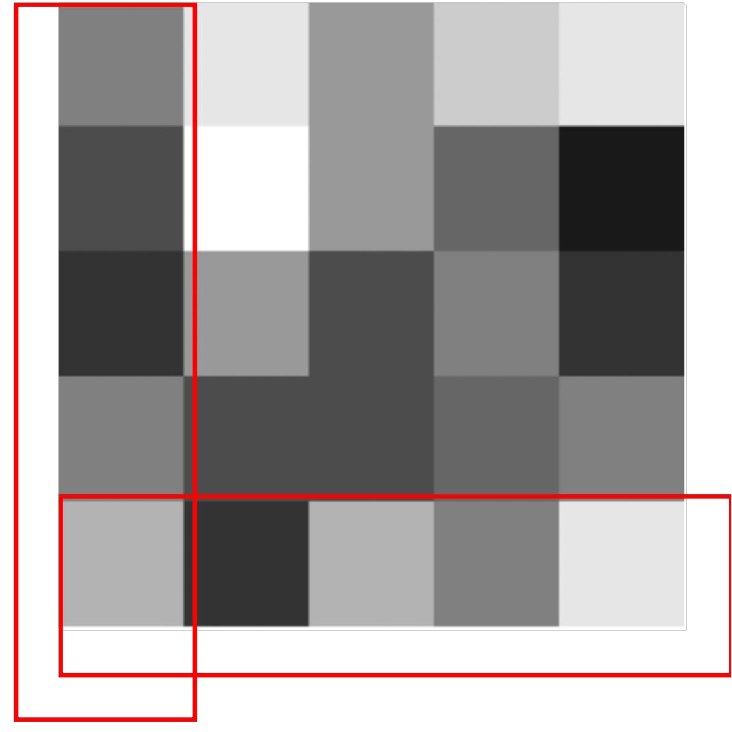}
    \includegraphics[width=0.143\textwidth]{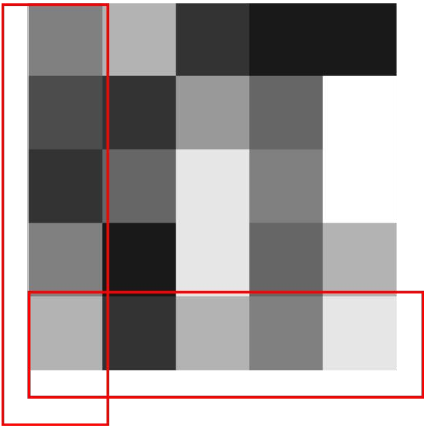}   
    \\
    \caption{The two images (b) and (c) share identical left and bottom boundaries and possess the same gradient magnitude $\left | g \right |$ (a), yet they are unmistakably distinct from each other.}
    \label{fig:example}   
\end{figure}
To illustrate this point, let's consider a basic $5\times5$ monochrome image with identical boundary values and grey value gradient magnitude which is shown in Figure~\ref{fig:example}. Despite these conditions, we can generate entirely different monochrome images that share the same gradient magnitude information. Even when more stringent conditions are applied, like knowing all boundary conditions, non-linear equations related to values still cannot determine all unknown values. In mathematical terms, reversing vectors based on their magnitudes is infeasible due to many-to-one mapping relationships, rendering this transformation inherently irreversible.
\paragraph{Color multiplicity.} We provide an example in Figure~\ref{fig:multi-mapping}, illustrating that multiple images may correlate with the same gradient amplitude information. In other words, a single color gradient module can correspond to numerous color gradient values and color values. This scenario implies that multiple images can share the same gradient magnitude information, making the retrieval of the original pixel values a formidable challenge. This lack of one-to-one correspondence adds a layer of complexity to privacy breach attempts.
\paragraph{Perceptual indistinction.} After gradient processing, the data becomes visually indistinguishable to human eyes. This transformation not only enhances privacy but also results in highly similar images for different identities. Learned Perceptual Image Patch Similarity (LPIPS) is a measure of perceptual similarity based on learning that is more in line with human perception. We use LPIPS to measure the similarity of images before and after privacy protection, as quantified in Table \ref{table:mytable}. This processing ensures that facial images are privacy-protected while maintaining their availability for geometric reconstruction tasks.

\begin{table}[ht]
  \centering
   \includegraphics[width=0.32\textwidth,height=0.18\textwidth]{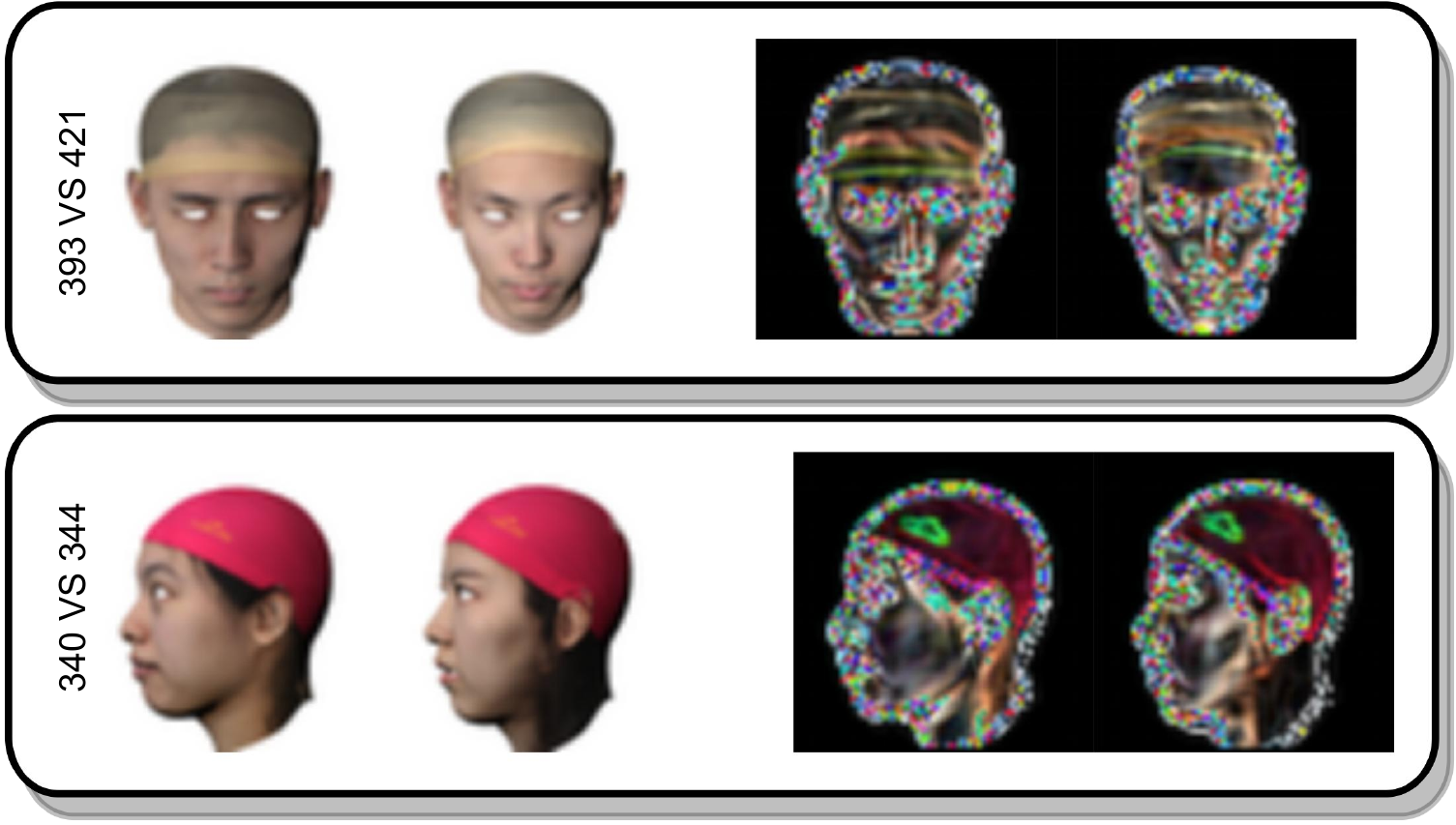}
  \resizebox{0.45\textwidth}{!}{
  \begin{tabular}{ccccccc}
    \toprule
     \rowcolor{lightgrey}& LPIPS  & 340 \textit{vs.} 344 & 375 \textit{vs.} 393 & 375 \textit{vs.} 421 & 393 \textit{vs.} 421 \\
    \midrule
     \rowcolor{lightgreen}& $\left \|\mathbf{g} \right \|_2$ images & 4.2778 & 4.2658 & 4.1261 & 4.1517 \\
    & RGB images & 4.6751 & 4.7295 & 4.6099 & 4.5353 \\
    \bottomrule
  \end{tabular}
  }\caption{LPIPS from different identities shows that the privacy-protected images are much more similar.}   
  \label{table:mytable}
\end{table}
\section{A broader evaluation}
\subsection{Robustness}
\begin{figure}\centering
    \centering
    \includegraphics[width=0.4\linewidth]{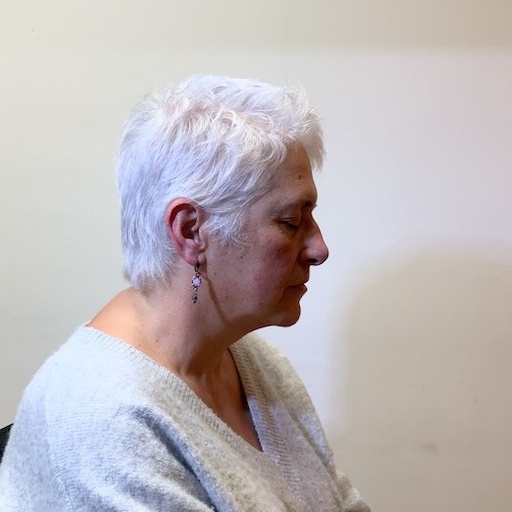}
    \includegraphics[width=0.4\linewidth]{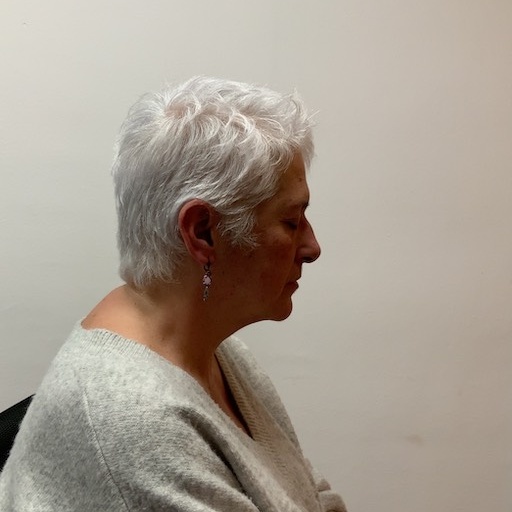}
    \caption{Our experiments are evaluated under different lighting conditions, especially the H3DS dataset.}
    \label{fig:lighting}
\end{figure}

These datasets we used in the paper cover a wide range of information, encompassing distinct lighting conditions (shown in Figure~\ref{fig:lighting}) and individuals of various races, skin tones, and ages. This diversity highlights the success of our experiments in achieving detailed geometric reconstruction across a spectrum of conditions.

We have introduced a highly flexible framework that currently utilizes a two-stage pipeline based on based on VolSDF~\cite{yariv2021volume} for reconstruction. To address additional geometric privacy scenarios, such as private meeting rooms or license plate reconstruction mentioned in our paper, we have adopted other neural reconstruction methods, such as 2DGS~\cite{huang20242d}, that are capable of managing these 3D scenes. Therefore, we encourage the exploration of neural rendering accelerators proposed in current research, such as~\cite{muller2022instant,huang20242d}, for potential optimizations in future work.
\subsection{Expressions}
\begin{figure} \centering
    \makebox[0.01\textwidth]{}
    \makebox[0.108\textwidth]{\small GT}
    \makebox[0.108\textwidth]{\small view 1}
    \makebox[0.108\textwidth]{\small view 2}
    \makebox[0.108\textwidth]{\small view 3}
    \\
    \raisebox{0.8\height}{\makebox[0.01\textwidth]{\rotatebox{90}{\makecell{\small angry}}}}
    \includegraphics[trim=1cm 0 3cm 0, clip,scale=1.5,width=0.108\textwidth]{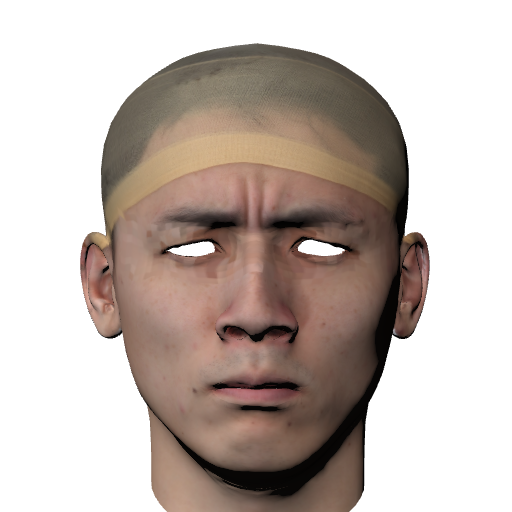}
    \includegraphics[trim=1cm 0 3cm 0, clip,scale=1.5,width=0.108\textwidth]{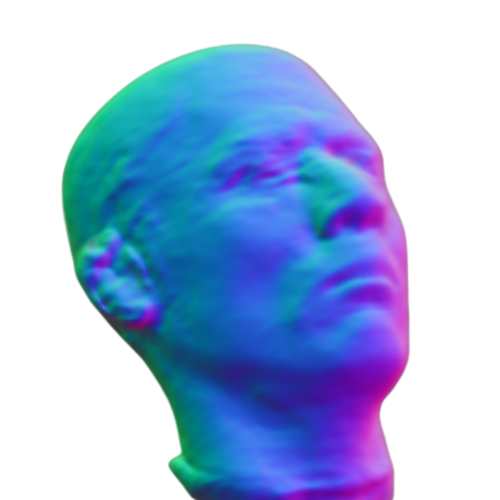}
    \includegraphics[trim=1cm 0 3cm 0, clip,scale=1.5,width=0.108\textwidth]{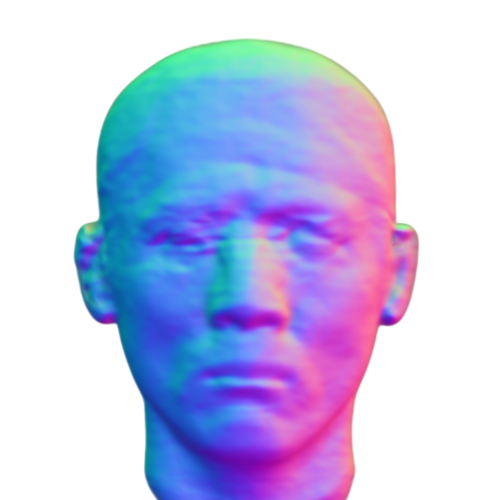}   
    \includegraphics[trim=1cm 0 3cm 0, clip,scale=1.5,width=0.108\textwidth]{image/evaluationnew/393_2-removebg-preview.png} 
    \\
    \raisebox{0.8\height}{\makebox[0.01\textwidth]{\rotatebox{90}{\makecell{\small shock}}}}
    \includegraphics[trim=1cm 0 3cm 0, clip,scale=1.5,width=0.108\textwidth]{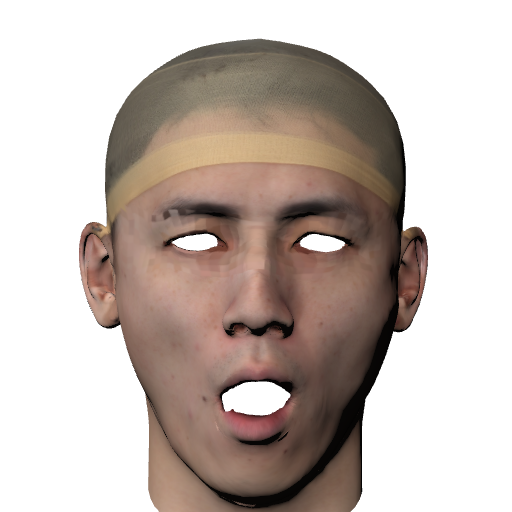}
    \includegraphics[trim=1cm 0 3cm 0, clip,scale=1.5,width=0.108\textwidth]{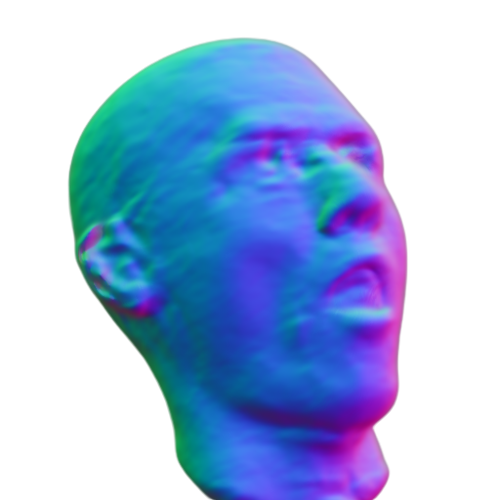}
    \includegraphics[trim=1cm 0 3cm 0, clip,scale=1.5,width=0.108\textwidth]{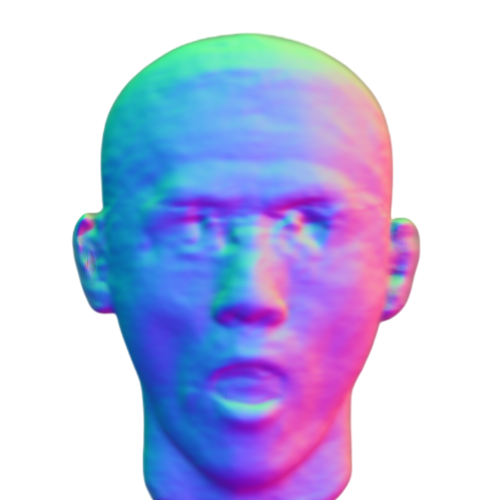}   
    \includegraphics[trim=1cm 0 3cm 0, clip,scale=1.5,width=0.108\textwidth]{image/evaluationnew/mouth2-removebg-preview.png} 
    \\
    \raisebox{0.8\height}{\makebox[0.01\textwidth]{\rotatebox{90}{\makecell{\small smile}}}}
    \includegraphics[trim=1cm 0 3cm 0, clip,scale=1.5,width=0.108\textwidth]{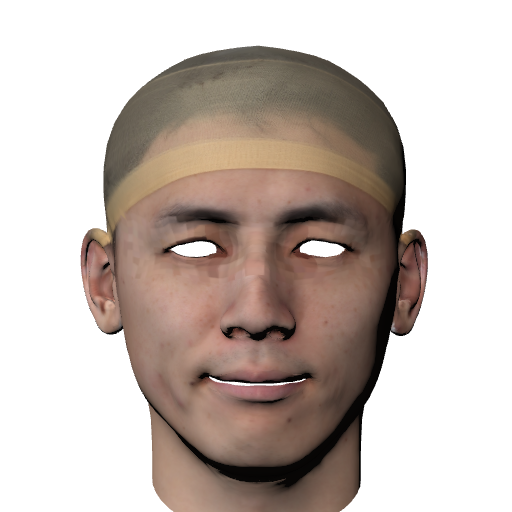}
    \includegraphics[trim=1cm 0 3cm 0, clip,scale=1.5,width=0.108\textwidth]{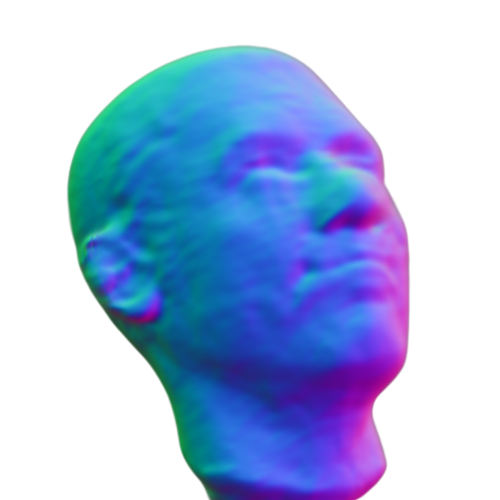}
    \includegraphics[trim=1cm 0 3cm 0, clip,scale=1.5,width=0.108\textwidth]{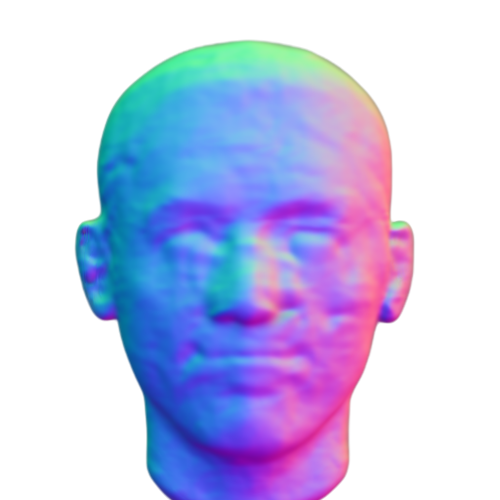} 
    \includegraphics[trim=1cm 0 3cm 0, clip,scale=1.5,width=0.108\textwidth]{image/evaluationnew/smile2-removebg-preview.png} 
    \\

    \caption{Our method excels in capturing variations in different expressions of an identity. As depicted in the figure, we provide three distinct expression examples for the same identity: `angry', `shocked', and `smile', encompassing facial frowns, mouth openings, and more. The detailed changes are well reflected in the geometric representations.} 
    \label{fig:emotion} 
    \end{figure}

Our model demonstrates robust performance across a wide range of facial expressions, showcasing its ability to effectively capture and represent subtle emotional nuances~\ref{fig:emotion}. This versatility highlights the model's potential for diverse applications, including advanced emotion recognition, and interactive AI systems. The expanded results underscore that our model is not only adept at handling common facial expressions but also offers detailed and nuanced representations, such as emotional states, which could be leveraged in innovative ways across various domains.

\section{Implementation}
\subsection{Datasets}
\begin{figure}
    \centering
    \includegraphics[width=0.25\textwidth]{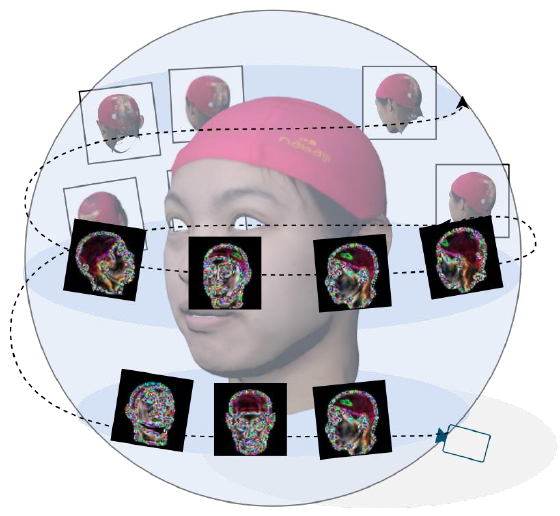}
    \caption{An example of facial image sensitivity classification from the FaceScape dataset.}
    \label{fig:wrap}
\end{figure}

\begin{figure} \centering
 
    \includegraphics[trim=1cm 0cm 3cm 1cm,  clip,scale=1.2,width=0.088\textwidth]{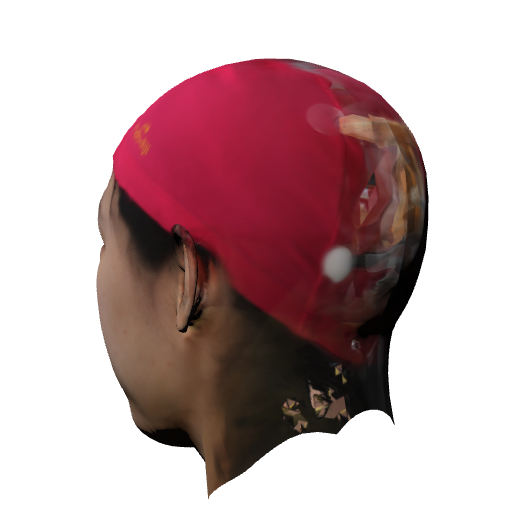}
    \includegraphics[trim=1cm 0cm 3cm 1cm,  clip,scale=1.2,width=0.088\textwidth]{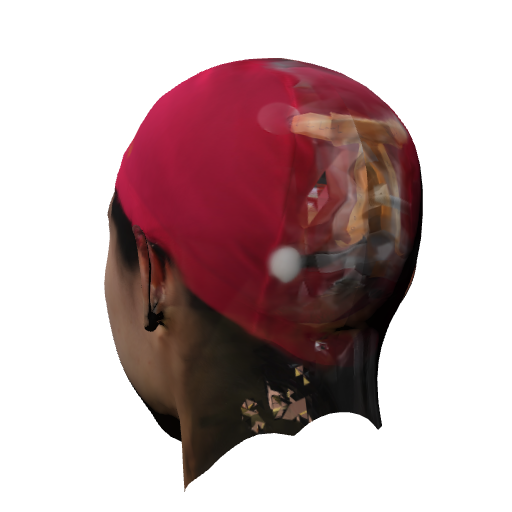}
    \includegraphics[trim=1cm 0cm 3cm 1cm,  clip,scale=1.2,width=0.088\textwidth]{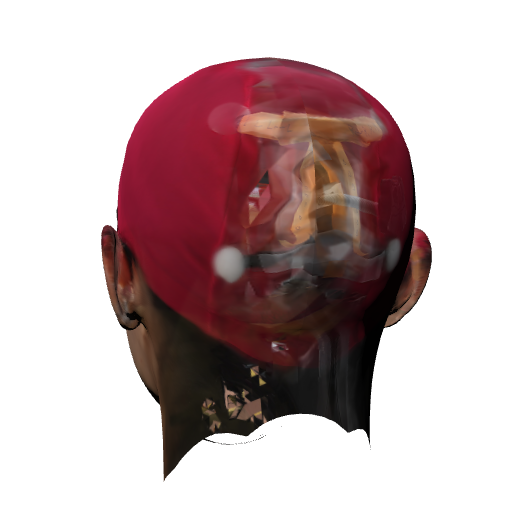}
    \includegraphics[trim=1cm 0cm 3cm 1cm, clip,scale=1.2,width=0.088\textwidth]{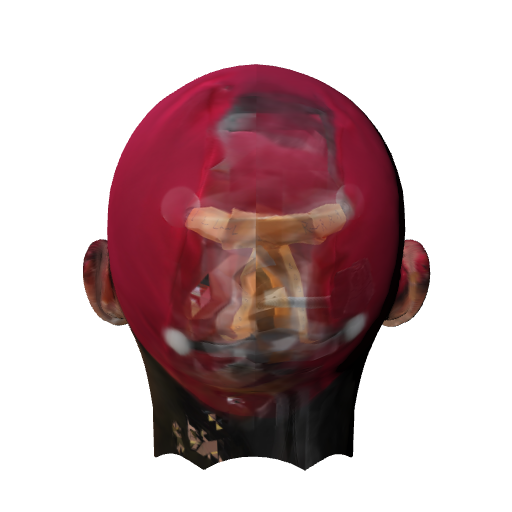}
    \includegraphics[trim=1cm 0cm 3cm 1cm, clip,scale=1,width=0.088\textwidth]{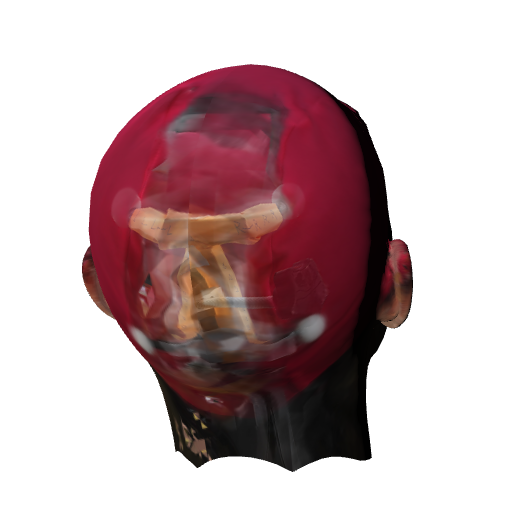 }
    \includegraphics[trim=1cm 0cm 3cm 1cm,  clip,scale=1.2,width=0.088\textwidth]{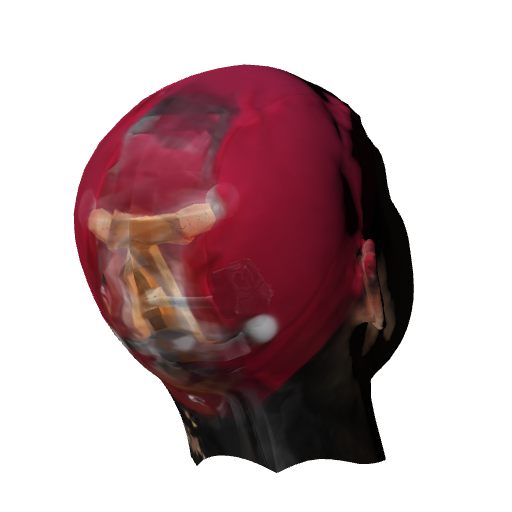}
    \includegraphics[trim=1cm 0cm 3cm 1cm,  clip,scale=1.2,width=0.088\textwidth]{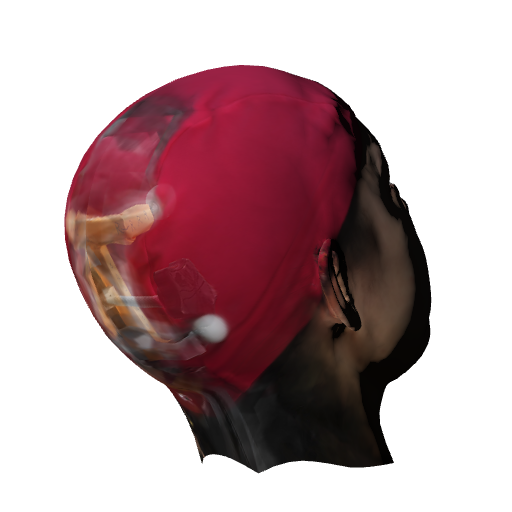}
    \includegraphics[trim=1cm 0cm 3cm 1cm,  clip,scale=1.2,width=0.088\textwidth]{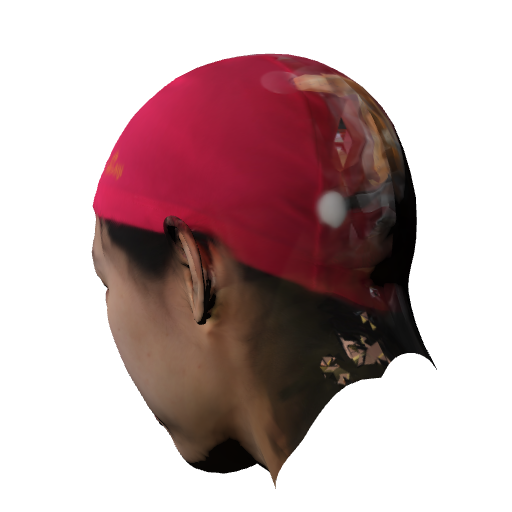}
    \includegraphics[trim=1cm 0cm 3cm 1cm, clip,scale=1.2,width=0.088\textwidth]{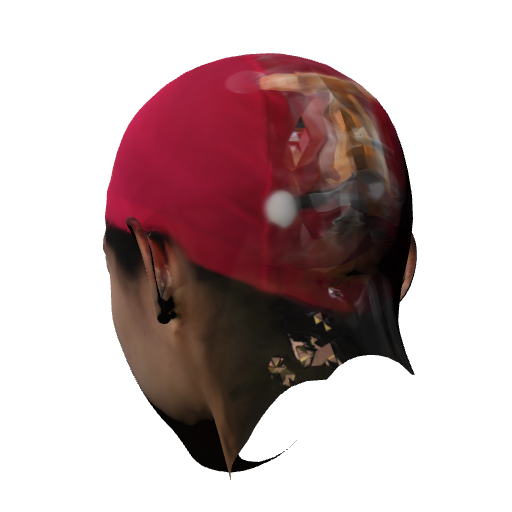}
    \includegraphics[trim=1cm 0cm 3cm 1cm, clip,scale=1,width=0.088\textwidth]{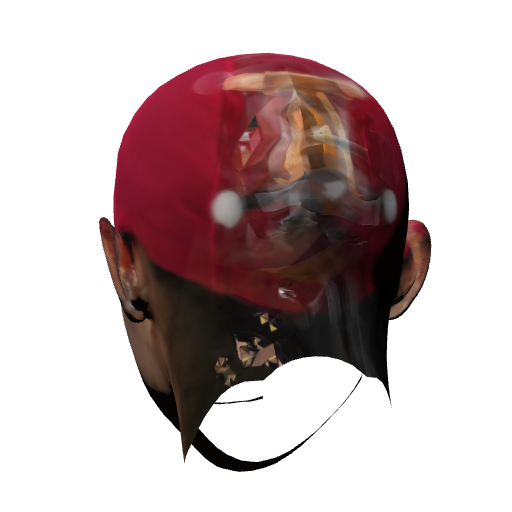 }
    \\   
    \includegraphics[width=0.088\textwidth]{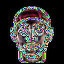}
    \includegraphics[width=0.088\textwidth]{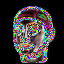}
    \includegraphics[width=0.088\textwidth]{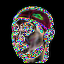}
    \includegraphics[width=0.088\textwidth]{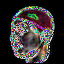}
    \includegraphics[width=0.088\textwidth]{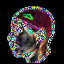 }
    \includegraphics[width=0.088\textwidth]{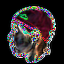}
    \includegraphics[width=0.088\textwidth]{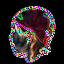}
    \includegraphics[width=0.088\textwidth]{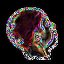}
    \includegraphics[width=0.088\textwidth]{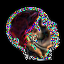}
    \includegraphics[width=0.088\textwidth]{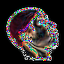 }
    \\
     \includegraphics[width=0.088\textwidth]{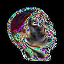}
    \includegraphics[width=0.088\textwidth]{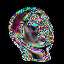}
    \includegraphics[width=0.088\textwidth]{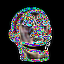}
    \includegraphics[width=0.088\textwidth]{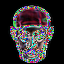}
    \includegraphics[width=0.088\textwidth]{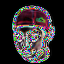 }    
     \includegraphics[width=0.088\textwidth]{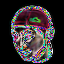}
    \includegraphics[width=0.088\textwidth]{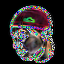}
    \includegraphics[width=0.088\textwidth]{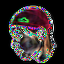}
    \includegraphics[width=0.088\textwidth]{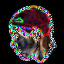}
    \includegraphics[width=0.088\textwidth]{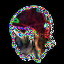 }
    \\    
        \caption{We have proposed a reconstruction method based on non-sensitive input. For neutral images that inherently do not contain sensitive information (Row 1,2), we retain the RGB images. For other images, we undergo protected processing before proceeding with subsequent operations(Row 3,4,5,6).} 
    \label{fig:supview}
\end{figure}

In the experimental section, for each group of data samples in the FaceScape dataset, we have 30 original RGB images taken from different camera perspectives, with the shooting perspective being consistent within each group. These images consist of two sets: one comprises neutral images directly usable for reconstruction, while the other contains sensitive information requiring protection (see Figure~\ref{fig:wrap}). In Figure \ref{fig:supview}, we showcase a complete set of image data used for the experiments from the FaceScape dataset, which intuitively represents our classification criteria.
\begin{figure}\centering

        \centering
        \raisebox{0.2\height}{\makebox[0.01\textwidth]{\rotatebox{90}{\makecell{\small Before}}}}
         \includegraphics[width=0.09\textwidth]{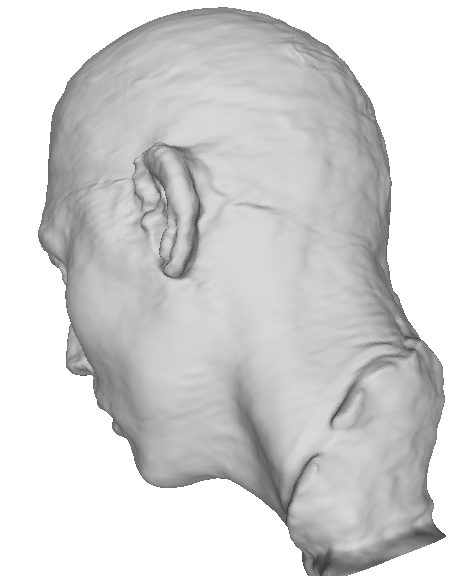}
         \raisebox{0.5\height}{\makebox[0.01\textwidth]{\rotatebox{90}{\makecell{\small After}}}}
        \includegraphics[width=0.088\textwidth]{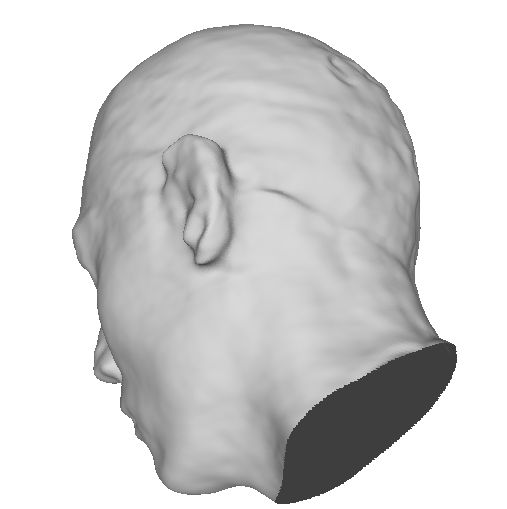}
        \includegraphics[width=0.088\textwidth]{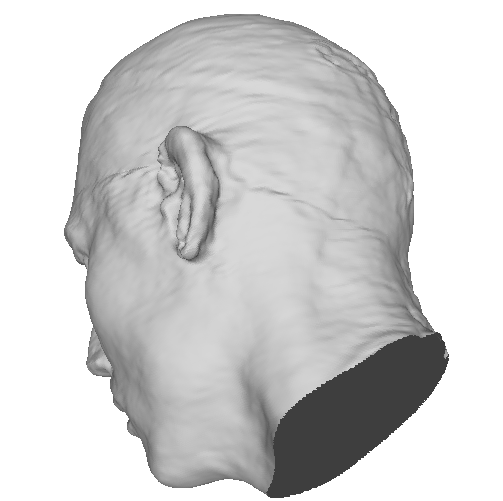}
\includegraphics[width=0.088\textwidth]{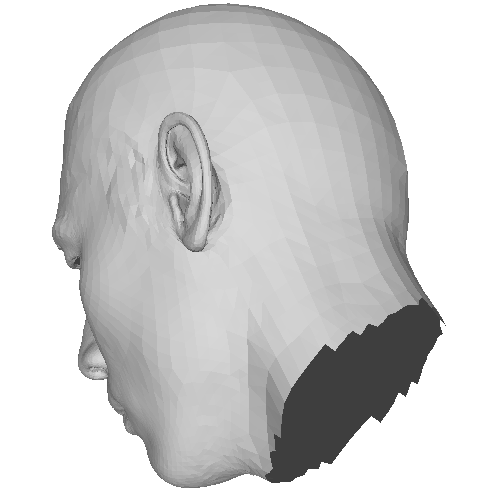}
\\
    \makebox[0.01\textwidth]{}
    \makebox[0.088\textwidth]{\small (a) VolSDF}
     \makebox[0.088\textwidth]{\small (b) Ours}
    \makebox[0.088\textwidth]{\small (c) VolSDF}
    \makebox[0.088\textwidth]{\small (d) GT}
    \caption{Cropping of the mesh is performed to ensure that the calculation of the Chamfer Distance (CD) values for all meshes is both reasonable and fair.}
    \label{fig:clip}
\end{figure}
Since in the H3DS dataset, the shooting perspectives may vary for each group of data samples, we employed a manual selection process with criteria similar to those mentioned in Figure~\ref{fig:supview}.

\subsection{Details on Training}
In Stage 1, we conducted training for 1,000 epochs to establish the foundation of facial geometry. During this phase, supervision was carried out exclusively using eikonal loss and RGB loss. At the conclusion of the first stage, the facial geometry lacked facial details.

Moving on to the second stage, we trained for an additional 500 epochs specifically focusing on reconstructing facial details. This phase involved using non-sensitive inputs and corresponding supervision to achieve a fine-tuning of the geometry, ensuring a more detailed representation of facial structures.
\subsection{Loss Setups}
In the first stage of the process, we employed supervision using only eikonal loss $\mathcal{L}_{\text{eik}}$ and RGB loss $\mathcal{L}_{\text{rgb}}$, with the total loss being calculated as:
\begin{equation}
  \mathcal{L}= \lambda_{1}\mathcal{L}_{\text{rgb}}+\lambda_{2}\mathcal{L}_{\text{eik}},  
\end{equation}
and we set $\lambda_{1}=1$, $\lambda_{2}=0.1$ to activate it.

In the second stage of the process, we introduced gradient loss $\mathcal{L}_{\text{grads}}$ to supervise the details. Additionally, to further enhance the reconstruction of geometry, we incorporated lip loss $\mathcal{L}_{\text{lip}}$ into the training regimen:
\begin{equation}
  \mathcal{L}= \lambda_{2}\mathcal{L}_{\text{eik}}+\lambda_{3}\mathcal{L}_{\text{lip}}+\lambda_{4}\mathcal{L}_{\text{grad}}, 
\end{equation}
and we set $\lambda_{2}=0.1$, $\lambda_{3}=3e^{-10}$, $\lambda_{4}=1$ to activate it.
\subsection{Details of Evaluations}

To ensure a fair comparison of the reconstruction results, we compared them with heads trained on a complete set of RGB images under the same conditions. For the FaceScape dataset, due to significant interference from the overall mesh of the head, we cropped facial mesh data, as illustrated in Figure \ref{fig:clip}, to ensure consistent size, encompassing all head geometry, for the purpose of comparison. For the H3DS dataset, we aligned the Ground Truth (GT) mesh with our obtained mesh using the alignment method provided by the H3DS dataset for a comprehensive comparative analysis.

\subsection{Template Acquisition}

We trained a set of head templates using an extensive dataset, with each template constructed from 10 unique identity heads. These templates were obtained through the method proposed by Xu et al. \cite{Xu_2023_ICCV}. The training process resulted in some template head geometry. Importantly, these templates are neutral and do not contain any sensitive information. The representation of a template facilitates its direct integration into the training process of the second stage, where protected inputs are utilized.
\subsection{Resolution Impact}

\begin{figure}\centering

        \centering
                 \includegraphics[width=0.12\textwidth]{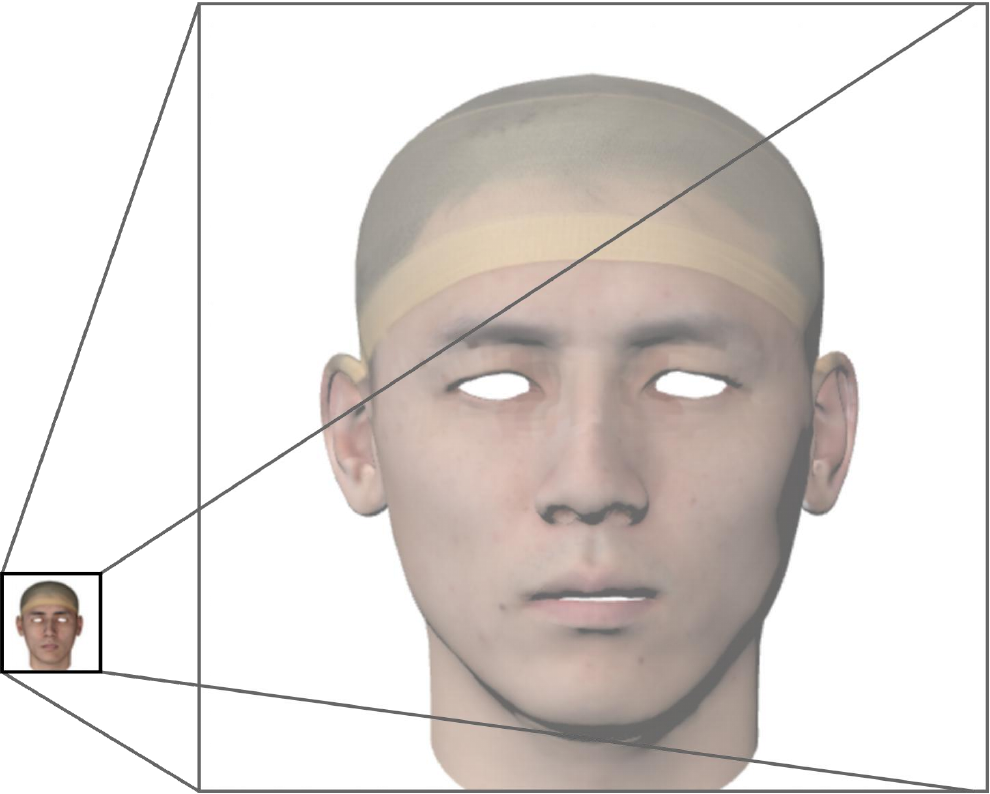}
        \includegraphics[width=0.12\textwidth]{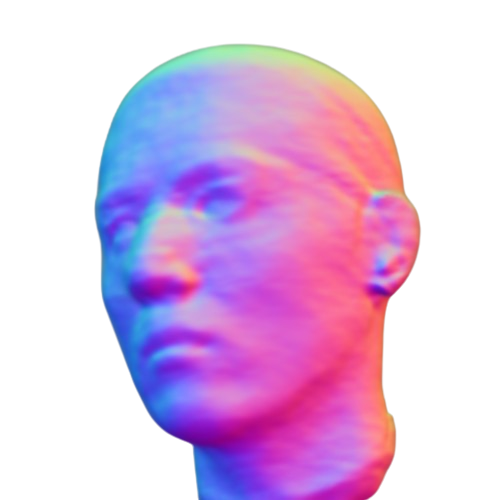}
        \includegraphics[width=0.12\textwidth]{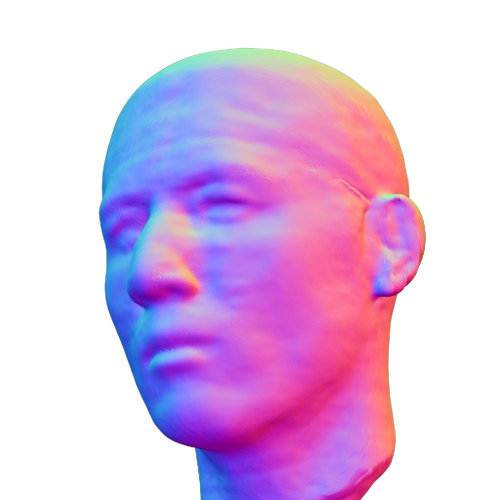}
        \\
    \makebox[0.01\textwidth]{}
    \makebox[0.12\textwidth]{\small (a) Down-sampling }
    \makebox[0.12\textwidth]{\small (b) 64-resolution}
    \makebox[0.12\textwidth]{\small (c) 512-resolution}
    \caption{(a) illustrates the downsampling of the original input images before further processing. (b) and (c) present the reconstruction results for supervised images at the 64-resolution level and 512-resolution level, respectively. It is evident that at higher resolutions, our method can restore more accurate geometric features.}
    \label{fig:resolution}
\end{figure}
It is important to note that, for the sake of sensitivity protection, we actively engage in the supervised reconstruction of all our images at a 64-resolution level. Despite the impact of resolution reduction on reconstruction quality, this effect remains unrelated to the robustness of our methodology. In Figure \ref{fig:resolution}, we depict the reconstruction outcomes obtained at both high resolution (512) and low resolution (64) through the utilization of non-sensitive input. This confirmation solidifies the understanding that the observed decline in geometric accuracy, which is within acceptable limits, is due to the supervision based on low-resolution images.
To ensure fair experiments, we maintain the same training epochs (1.5k) and employ $512\times512$ resolution matching cubes to extract the final mesh. The mesh evaluation is conducted by appropriately cropping it to the size of the provided GT mesh, ensuring a fair assessment of facial geometric reconstruction accuracy.


\section{Acknowledgments}
This work was supported in part by the Ministry of Education, Singapore, under its Academic Research Fund Grants (MOE-T2EP20220-0005 \& RT19/22) and the RIE2020 Industry Alignment Fund–Industry Collaboration Projects (IAF-ICP) Funding Initiative, as well as cash and in-kind contribution from the industry partner(s).

\bibliography{aaai25}

\end{document}